\documentclass{egpubl}
\usepackage{eg2019}
\usepackage{amsmath}

\ConferencePaper        
 \electronicVersion 

\ifpdf \usepackage[pdftex]{graphicx} \pdfcompresslevel=9
\else \usepackage[dvips]{graphicx} \fi

\PrintedOrElectronic

\usepackage{t1enc,dfadobe}

\usepackage{egweblnk}
\usepackage{cite}

\title[Clear Skies Ahead]%
      {Clear Skies Ahead: Towards Real-Time Automatic Sky Replacement in Video}

\author[T. Halperin \& H. Cain \& O. Bibi \& M. Werman]
{\parbox{\textwidth}{\centering Tavi Halperin$^{1,2}$\orcid {0000-0001-9288-5392},
        Harel Cain$^{1}$, 
        Ofir Bibi$^{1}$ and
        Michael Werman$^{2}$
        }
        \\
{\parbox{\textwidth}{\centering $^1$Lightricks\\
         $^2$The Hebrew University of Jerusalem, Israel
       }
}
}

\begin{document}

\teaser{
\centering 
{\includegraphics[width=1.0\linewidth]{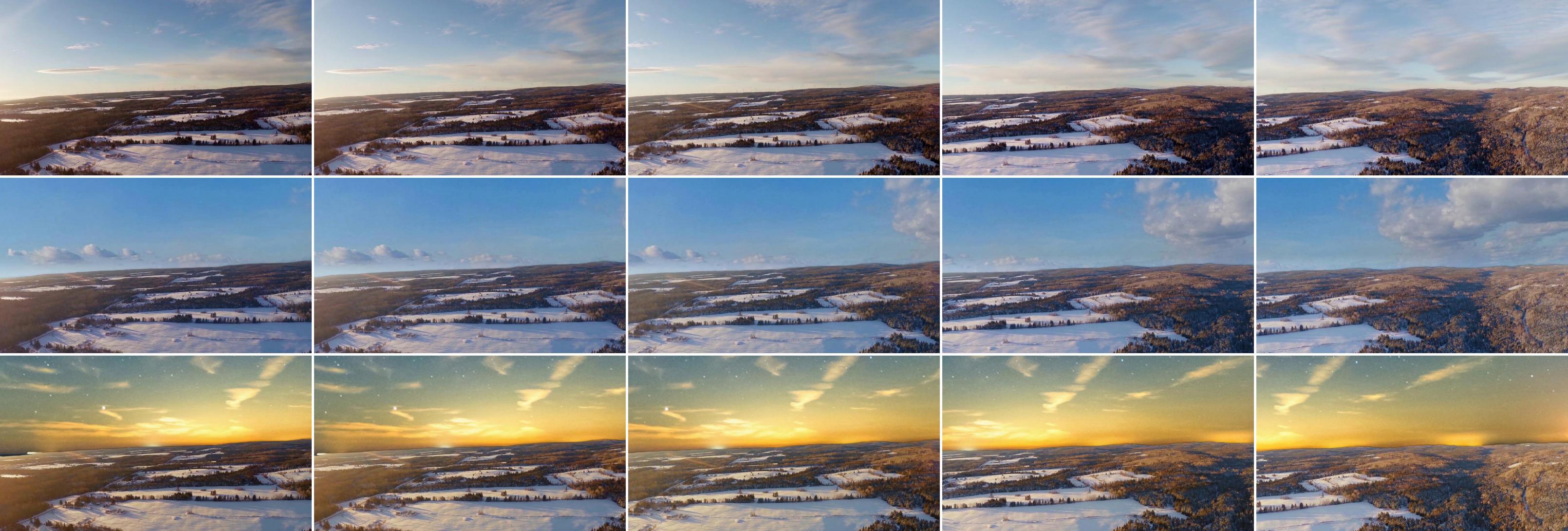}}

\caption {\textbf{Top:} sampled frames of an input video. \textbf{Middle and bottom:} sky  replaced   from 
other videos which were generated from panoramic images according to the motion in the upper row.
}
\label{fig:teaserfigure}
}

\maketitle

\newcommand{\baseVid}{\textit{base }}
\newcommand{\coverVid}{\textit{sky }}

\begin{abstract}
Digital videos such as those captured by a smartphone often exhibit exposure inconsistencies, a poorly exposed sky, or simply suffer from an uninteresting or plain looking sky.
Professionals may edit these videos using advanced and time-consuming tools unavailable to most users, to replace the sky with a more expressive or imaginative sky. 
In this work, we propose an algorithm for automatic replacement of the sky region in a video with a different sky, providing nonprofessional users with a simple yet efficient tool to seamlessly replace the sky. 
The method is fast, achieving close to real-time performance on mobile devices and the user's involvement can remain as limited as simply selecting the replacement sky.

\ccsdesc[500]{Computing methodologies~Computational photography}
\ccsdesc[300]{Computing methodologies~Image processing}

\printccsdesc   
\end{abstract}

\newcommand{\R}[1]{{\color[rgb]{1,0,1} {\Large #1}}}

\section{Introduction}

Sky in outdoor videos poses a challenge for photographers. The location for shooting a video may be chosen carefully, yet the sky which often covers a large portion of the frame is subject to uncontrollable weather and lighting conditions.
To fix this, methods for sky segmentation and replacement in still images have been studied \cite{Tsai2016SkyIN,2017arXiv171209161L}. 
We build upon these works and extend them to video. Simply applying sky replacement frame by frame rarely works,  even if inefficiently, and lacks components handling camera motion. When tackling the full scope of sky replacement in video we encounter many issues in need of resolution. These issues include algorithmic runtime efficiency, segmentation temporal consistency, lighting compensation, and camera motion matching.

In this paper we focus on videos taken by a handheld device. 
We assume the sky is infinitely far away. 
Thus, pure translation of the camera (i.e no rotation) will not displace sky pixels in the image, and rotating the camera results in a homography between images. It was suggested in \cite{szeliski2010computer} to use the infinite homography $H_{\infty}$, 
to model the transformation between images taken roughly from the same location. This homography is computed between points far away from the camera. Since in a video taken by a handheld camera its translation is small relative to the rotation, $H_{\infty}$ is a good approximation of the sky's motion.

The \textit{replacement} sky, taken from a spherical $360^{\circ}\times180^{\circ}$ image (or video), which we refer to as the \coverVid image, is used to replace the sky in the \baseVid video.

Working with video instead of a single image provides the challenge of matching geometric properties of the sky sampled from the \coverVid image to those of the \baseVid video. Sampling a perspective projection requires determining the field of view (FOV) of the \baseVid video since we need equal FOVs for the  videos so that the motions match. The FOV can be calculated from the focal length, which is commonly contained in the EXIF tags of images. For videos, however, the focal length is generally not provided, and although calibrating a rotating camera is a well studied problem \cite{hartley1994self}, when the camera is also slightly translating in addition to its rotation, errors in focal length estimation add up. We use a slightly more robust calibration method to better fit this setup.

Another drawback of applying existing single image sky replacement approaches to videos, is the running time. For example, a running time of 12 seconds per frame as in  \cite{Tsai2016SkyIN} may seem reasonable for a single image, but when performing the task on a video with hundreds or thousands of frames, efficiency becomes essential. Our work is developed with efficiency in mind and we adapt the components of our framework to achieve close to real-time performance.

\section{Related Work}
Sky replacement depends on sky segmentation, recovering camera rotation and focal length parameters, and matching photometric properties between the two sources. We review the most relevant work in these areas.

\subsection{Semantic Segmentation}
Semantic segmentation has seen tremendous advances in recent years \cite{Shelhamer2015FullyCN,Chen2017DeepLabSI,Zhao2017PyramidSP,Zhao2017ICNetFR}.
Our goal is to provide a semantic segmentation model trained to detect the sky region in arbitrary, unconstrained 'in the wild' videos, which is consistent under changing conditions which are common to video such as camera motion and lighting variations. 
Few if any annotated datasets for semantic segmentation of videos are truly 'in the wild'. Most video segmentation datasets are very constrained in their domain, for example CamVid \cite{Brostow2009SemanticOC} and Cityscapes \cite{Cordts2016TheCD}, which are limited to videos captured in urban landscapes from a driving car. We thus preferred to train a sky segmentation model on still image datasets, augmenting these datasets to enforce segmentation consistency, and then apply it to semantic segmentation of video. Historically, datasets such as MS-COCO \cite{Lin2014MicrosoftCC} focus on 'things' such as salient objects and not on 'stuff' such as major scene background components. In recent years new datasets that also include 'stuff' categories were collected. We  used three publicly available image datasets annotated for semantic segmentation which contain a sky class: Pascal-Context \cite{Mottaghi2014TheRO}, COCO-Stuff \cite{Caesar2016COCOStuffTA} and ADE20K \cite{Zhou2017ScenePT}. 

\subsection{Image Composition}
The na\"ive approach to cut-and-paste segmented areas from different images will usually result in unnatural looking composites as source images will likely differ in lighting conditions \cite{lalonde2007using}. Several image composition techniques were proposed to assess and improve realism of composite images \cite{tsai2017deep, sunkavalli2010multi, zhu2015learning, xue2012understanding}. They focus on transferring colors of one of the images so their statistics match the color statistics of the other image. 
The color transfer parameters in \cite{zhu2015learning} are obtained by optimizing an affine color transform so that the composite image scores high 'objective' realism measure. The score is obtained by feeding the composite image to a CNN trained to distinguish composite images from real photographs. We use this pretrained CNN to automatically compare realism of videos with replaced sky vs. their original counterparts.

\subsection{Single Image Sky Replacement} 
A special case of image composition is sky replacement. Following sky segmentation, Tao et al. \cite{tao2009skyfinder} provide an attribute based search for an adequate \coverVid image. Tsai et al. \cite{Tsai2016SkyIN} use FCNN segmentation both to segment the sky and to retrieve candidates from which to transfer sky, based on semantic layout similarity. They also extend the rather simple color transfer technique employed in \cite{tao2009skyfinder}. Both approaches have natural looking results. We share some of the building blocks with \cite{Tsai2016SkyIN}, focusing on the special challenges in video. 

\subsection{Camera Motion Recovery} 
In addition to an image composition technique, composing \textit{video} requires an accurate camera motion estimation. This has been an active area of research for over a century. We will just mention the few most relevant works. Intuitively, since we assume the camera is under the same 'skydome' for the entire clip, we are only interested in the relative camera rotation independent of the translation. At first glance, the work by Kneip et al. \cite{kneip2012finding} may seem to match our needs perfectly. However, in our experiments, this method did not produce exact enough results, requires the intrinsic camera parameters to be known in advance, and is not suited for RANSAC.

There is a myriad of approaches for structure from motion (SfM), also known as simultaneous localization and mapping (SLAM) (e.g. \cite{vijayanarasimhan2017sfm, klein2007parallel, mur2015orb, engel2017direct, engel2014lsd}), which may be utilized to recover camera motion. These approaches generally require significant camera translation for high accuracy. Model selection techniques \cite{torr1998maintaining, torr1997assessment} were suggested to distinguish pure camera rotation from general displacement. We focus on handheld cameras moving freely in space, where the majority of the motion is rotational and with only a small translation compared to the scale of the scene. For such scenes it is often more accurate to model the motion  as a projective transformation \cite{szeliski2010computer}.

\begin{figure*}
\centering
  \setlength{\tabcolsep}{3pt}
{\includegraphics[width=1.0\linewidth]{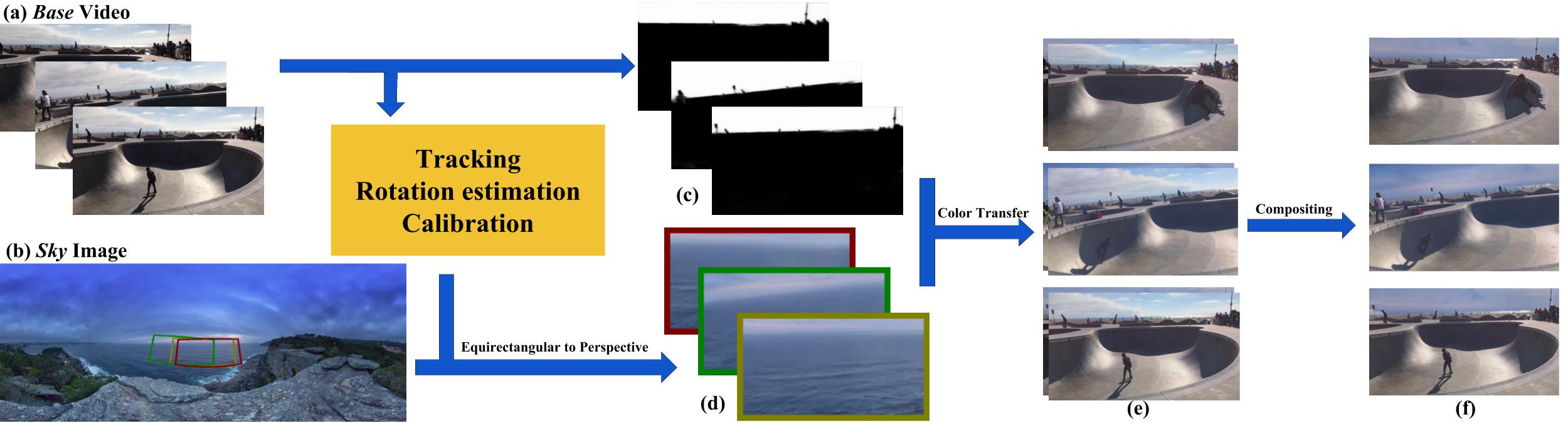}} 
\caption{Method Overview. First, the \baseVid video (a) is fed to a sky segmentation network (c) and tracking is performed, followed by calibration and motion estimation. Based on this estimation, a video mimicking the motion (d) is generated from the \coverVid image (b). Tonal adjustments are performed to increase realism of the output (e) and finally the two videos are blended according to the mask(f).}
\label{fig:pipeline}
\end{figure*}

\subsection{Photometric Calibration}
Another difference between single image and video composition is exposure changes between consecutive frames of video. Exposure variations of combined videos should be compatible, thus exposure change in one video should be applied to the other before combining them. To recover exposure variation, Goldman and Jiun-Hung Chen \cite{goldman2005vignette} developed an optimization function to simultaneously estimate the camera response curve (CRC), variation in exposure, and vignetting. To improve efficiency, \cite{bergmann2018} optimize the function with an analytic Jacobian. We improve it further by limiting the temporal span over which the CRC and vignetting are computed for faster convergence.

\section{Algorithmic Overview}
Sky replacement in video depends on a number of techniques; sky segmentation with temporal consistency, focal length estimation, computation of camera rotation parameters between consecutive frames, computing photometric properties, color transfer and compositing of the \coverVid image into the \baseVid video. In order not to compute everything in each frame information flows between frames based on tracked feature points.

The first steps, sky segmentation and tracking, can be carried out in parallel. The tracked points are then used to estimate camera rotation, focal length, vignetting and exposure changes. Then, a video that mimics the motion of the \baseVid video is created from the \coverVid image. Then finally the \baseVid video is color graded in order to allow for more natural looking compositions with the sky region of the created \coverVid video as they are composited. An outline of the  process is illustrated in Figure \ref{fig:pipeline}, and detailed in the following sections.

\section{Sky Segmentation}
Precise and temporally consistent semantic segmentation of the sky  in the \baseVid video is a prerequisite for any sky replacement operation. For our task, we are also concerned with the computational cost of this step, as users will expect sky replacement in videos not to take  much more time than playing the video even in off-line processing. More importantly, real time is compulsory for augmented reality applications.

\subsection{Datasets and Data Organization}
We used images from Pascal-Context \cite{Mottaghi2014TheRO}, COCO-Stuff \cite{Caesar2016COCOStuffTA} and ADE20K \cite{Zhou2017ScenePT} where the pixels from the relevant images from the three
 datasets were merged into a simple two-class division of sky and non-sky, and 'cloud' was considered sky. The network is trained to predict a two-class score for every pixel using a Softmax activation at the last layer. We  collected more than 14,000 images with a ground-truth mask for the sky region and partitioned  this  dataset into   train, validation and test sets comprised of 64\%,16\%,20\% of the dataset respectively. 

To deal with the relative lack of semi-clouded, high-contrast skies in the dataset, we augmented the original images by pasting various forms of clouds (represented as images with a transparent background) in random locations within the ground-truth sky area, increasing the ability of the network to identify high-contrast clouds on clear skies. To demonstrate the improvement this data augmentation scheme gave, we trained two identical models on exactly the same dataset up to the addition of the pasted clouds to the sky area. When they were later tested on held-out images from COCO-Stuff \cite{Caesar2016COCOStuffTA} which contain the cloud category in their annotations (thus they contain natural, non-pasted clouds), we observed an average IoU of 89.4\% for the model trained with the pasted clouds, compared to 88.4\% of the model trained without them. The first model achieved 68.2\% of images with an IoU higher than 90\%, considerably higher than the second model which achieved 63.5\% of such images.

Another augmentation process we used involved combinations of geometric and tonal transformations, applied randomly during training, with different parameters for each image in each training epoch. Geometric transformations included vertical flip, horizontal flip, small rotations, random crops and perspective transformations. Tonal transformations included brightness and contrast changes, conversion to grayscale, addition of white Gaussian noise, and changes of hue and saturation.

\begin{figure}
\centering
{\includegraphics[width=1.0\linewidth]{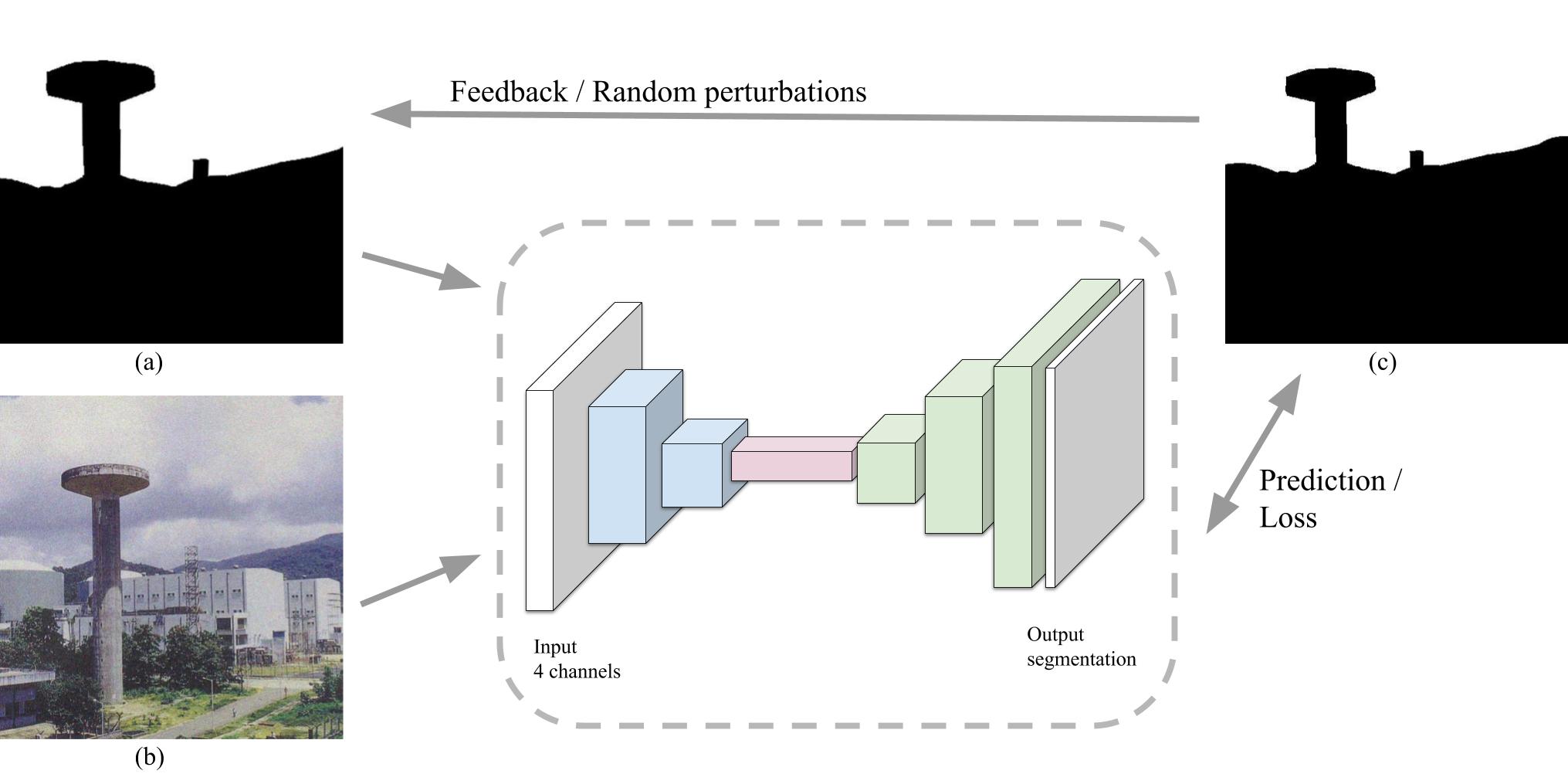}} 
\caption{Network training and inference procedure. We train on a dataset of individual images and simulate effects of video. During \emph{training}, the ground truth mask (c) is perturbed by slight piecewise affine transformations and added noise. The perturbed mask (a) is concatenated with the input image (b) and fed to the network while the unperturbed mask is used for loss. During \emph{inference}, the predicted mask of frame $i$ is fed back into the network for predicting the next frame's mask.}
\label{fig:network}
\end{figure}

\subsection{Network Architecture}
We designed a relatively small segmentation model, inspired by various network architectures that have been shown to be beneficial for semantic segmentation tasks. Our network contains: three feedforward blocks, each including: a convolution layer with 3x3 kernels, a maxpooling layer and a batch normalization layer; a series of residual blocks in the bottleneck stage, inspired by ResNet and similar architectures, but with the full pre-activation design proposed in \cite{He2016IdentityMI}, and three top-down SharpMask \cite{Pinheiro2016LearningTR} blocks with skip connections, which scale the spatial dimension of the result back to that of the input, and help preserve fine detail. The Sharpmask blocks are then followed by two "fully connected" 1x1 convolutional layers representing the final decision per pixel. 


\subsection{Feedback channel for temporal consistency}
To enhance the network's temporal consistency over consecutive video frames, we employ a feedback channel in which the previous video frame's predicted segmentation mask is fed as a fourth channel in the input tensor of the network in addition to the three RGB channels of the current frame. This channel serves as a reliable estimate of the current frame's correct segmentation mask, such that the network has only to learn how to adapt it to the changes between the current frame and the previous one due to scene motion and camera motion.
The main challenge with this approach is how to train the network on annotated image datasets (for the lack of densely annotated relevant video datasets), so that the network will learn not to ignore the fourth channel in its input, but not rely on it too much when there is a lot of motion between consecutive frames. To do so, during training time the fourth channel is populated with one of the following (on a random basis): most of the time, a small random piecewise-affine transformation of the image's ground truth mask is used (this serves as the proxy for the previous frame's segmentation mask during the online inference phase); the rest of the time, we use either an all-black mask, a random noisy mask with low-passed white noise or the slightly perturbed ground truth mask combined with such a noise pattern.

The network architecture is illustrated in Figure \ref{fig:network}.
\section{Estimating Camera Motion}

The camera motion computation is based on tracked points between frames.
In order to adhere to our rotation only motion model, we exclusively track far away objects. Ideally, these should be sky pixels, as we already have them segmented. However, sky is hard to track, with few or even no 'good features', and even when sky pixels can be reliably tracked they may only cover small areas in some of the frames, resulting in inaccurate motion estimation.

To still get a reliable motion estimation, we make use of the observation that the motion a handheld camera undergoes in an outdoor environment is often best modeled by a purely rotational constraint \cite{szeliski2010computer}, which allows us to track non-sky pixels as well, and still get an accurate motion estimation.
We use the KLT pyramidal tracker \cite{shi1994good} to detect and track feature points. We experimented with other descriptor based trackers but did not get an improvement in accuracy, only a degradation in efficiency. We divide the frame into cells and detect 'good features' in each of them. This improves motion estimation as the homography will tend to be computed over the largest possible field of view, and also improves vignetting estimation which benefits from dense sampling.

To ensure at each frame that the remaining tracked trajectories are  spread out, 
we  compute the SVD  of the  point locations. In  subsequent frames,  if the ratio between any of the singular values and the initial ones falls below a threshold, we set the previous frame (in which this threshold was not yet crossed) as a keyframe, and we detect new points to track forward.

In subsequent frames $i$ we evaluate the forward-backward error \cite{kalal2010forward} to pre-filter outlier tracks, and compute the projective transformation $H_{i}$ to the last keyframe using RANSAC.

We concatenate homographies until there exists a homography between every frame to the first one. They are used to calibrate the camera and to compute rotations.

\subsection{Camera Calibration}

We assume  constant intrinsic camera  parameters  $K$  throughout the entire video. The homography $H$ is decomposed as
\begin{equation}
H=KRK^{-1}
\end{equation}
where $R$ is a rotation matrix and $K$  an upper triangular matrix.  We set the image  origin at the center of the image and assume that it is the center of projection. We further assume zero skew and a single focal length in  $x$ and $y$ directions. $K$ is thus
\begin{equation}
\label{k_definition}
K=
\begin{bmatrix}
    f & 0 & 0  \\
    0 & f & 0  \\
    0 & 0 & 1
\end{bmatrix}
\end{equation}
with the focal length $f$,  the only unknown. A pure rotational motion implies that $H$ is an orthogonal conjugate matrix. That is, after normalizing $H$ so that $det(H) = 1$ its eigenvalues are $\{1, e^{i\theta}, e^{-i\theta}\}$. This is known as the modulus constraint \cite{pollefeysmodulus} and is used to calibrate a rotating camera \cite{hartley1994self}. In the context of motion model selection this approach is frowned upon in \cite{torr1998maintaining} for two reasons: (1) If the two images were taken with different intrinsic parameters (e.g due to autofocus) it would fail; (2) In planar motion the homography is a conjugate rotation, even though the true motion includes translation of the camera. However, in outdoor environment (1) is not likely, and (2) actually works to our favor, since the translation  is eliminated for free. 

Assuming pure rotation the following equation holds \cite{szeliski2010computer}
\begin{equation}
f^{2} = \frac{h_{12}^2-h_{02}^2}{h_{00}^2+h_{01}^2-h_{10}^2-h_{11}^2}
\end{equation}
where $h_{ij}$ is the element of $H$ in row $i$ and column $j$. This solution for $f$, as well as the more complicated one which solves for all intrinsic parameters presented in \cite{hartley2003multiple}, assumes that $H$ is an orthogonal conjugate matrix. We relax this constraint. Denote the eigen-decomposition $H = VDV^{-1}$, 
where the columns of $V$ are the eigenvectors of $H$, corresponding to the eigenvalues which are on the diagonal of $D$. If the motion has non zero translation the eigenvalues may deviate from unity modulus. Substituting the RQ decomposition $V=KU$ where $K$ is an upper-triangular matrix and $U$  hermitian, yields the following equation
\begin{equation}
H=KUDU^{-1}K^{-1}=K\tilde{R}K^{-1}
\end{equation}
Enforcing positive values on the main diagonal of $K$ and normalizing so that $k_{33}=1$ removes all ambiguities and results in a unique decomposition. The relaxation comes from the fact that $\tilde{R}$ need not be orthogonal.

Although all intrinsic parameters are computed as the upper right triangular matrix $K$, they are not equally reliable. It was suggested in \cite{hartley2003multiple} to use at least 3 images acquired by a camera rotating in different directions to reliably recover $K$, because for example for a panning camera (rotating around Y axis) it is impossible to recover the focal length along the rotation axis. We, however, assume equal focal length along both axes, and therefore extract only $k_{11}$ from $K$ and set it as the focal length. To improve accuracy and avoid the panning ambiguity we rectify the feature locations used to compute $H$ by rotating them in pixel space such that the axis of the 3D rotation between the cameras coincides with the $y$ axis. The axis of rotation between cameras is the eigenvector in $V$ corresponding to the real eigenvalue in $D$. This ensures accurate focal length along the $x$ axis and we use the same value for focal length along the $y$ axis.

Finally, to robustify the estimation, the focal length is taken to be the median over those computed from homographies corresponding to large rotation angles ($\theta$ from the eigenvalue) and motion closest to pure rotation. To measure the deviation of the motion from a pure rotation we use deviations of the eigenvalues from unit magnitude. To recover camera rotation, once calibration is fixed we transform the image corners $c_2=Hc_1$ and compute the orthogonal matrix $R$ which minimizes $\|RK^{-1}c_1-K^{-1}c_2\|_2$.

\begin{figure}
\centering
  \setlength{\tabcolsep}{3pt}
  \begin{tabular}{cc}
{\includegraphics[width=0.5\linewidth]{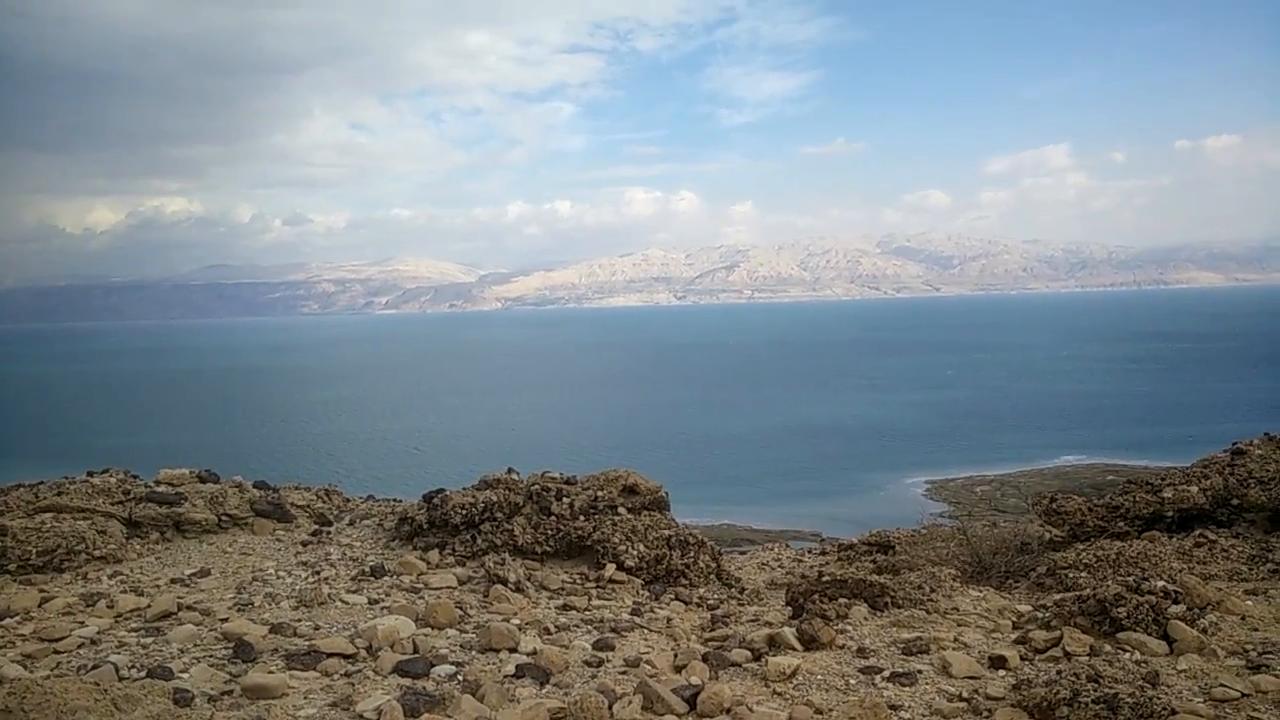}} &
{\includegraphics[width=0.5\linewidth]{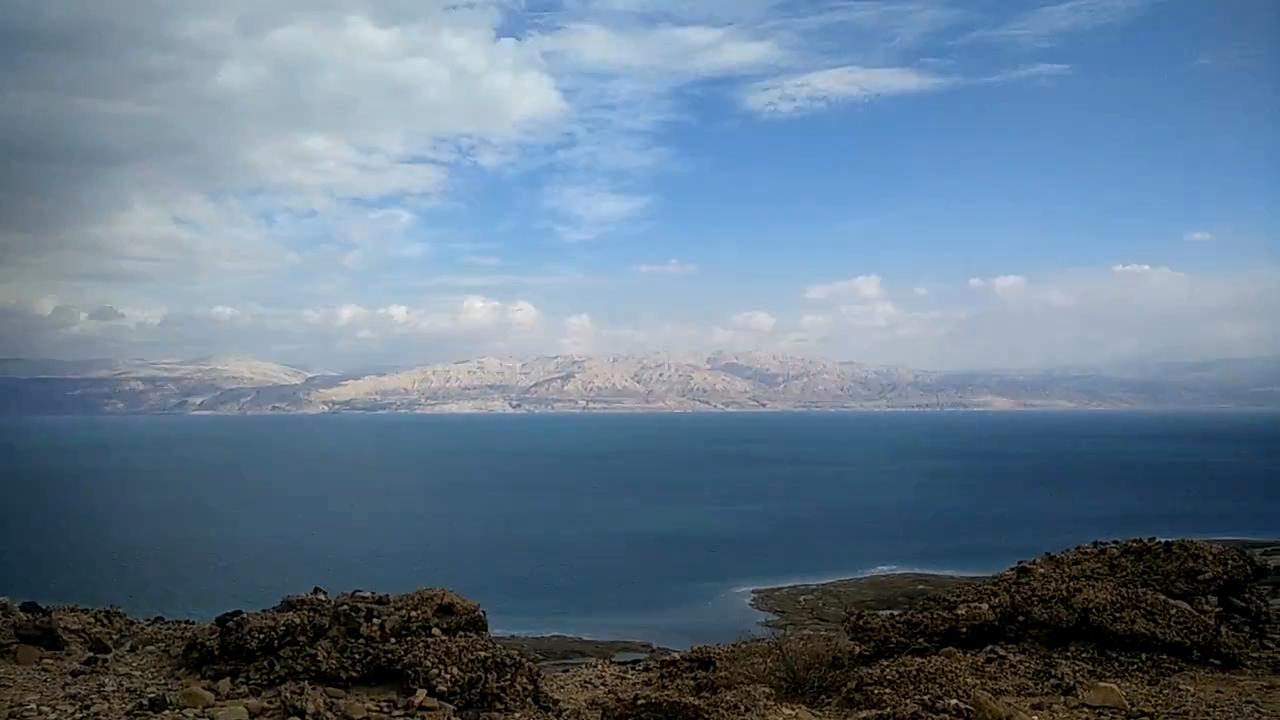}} \\
{\includegraphics[width=0.5\linewidth]{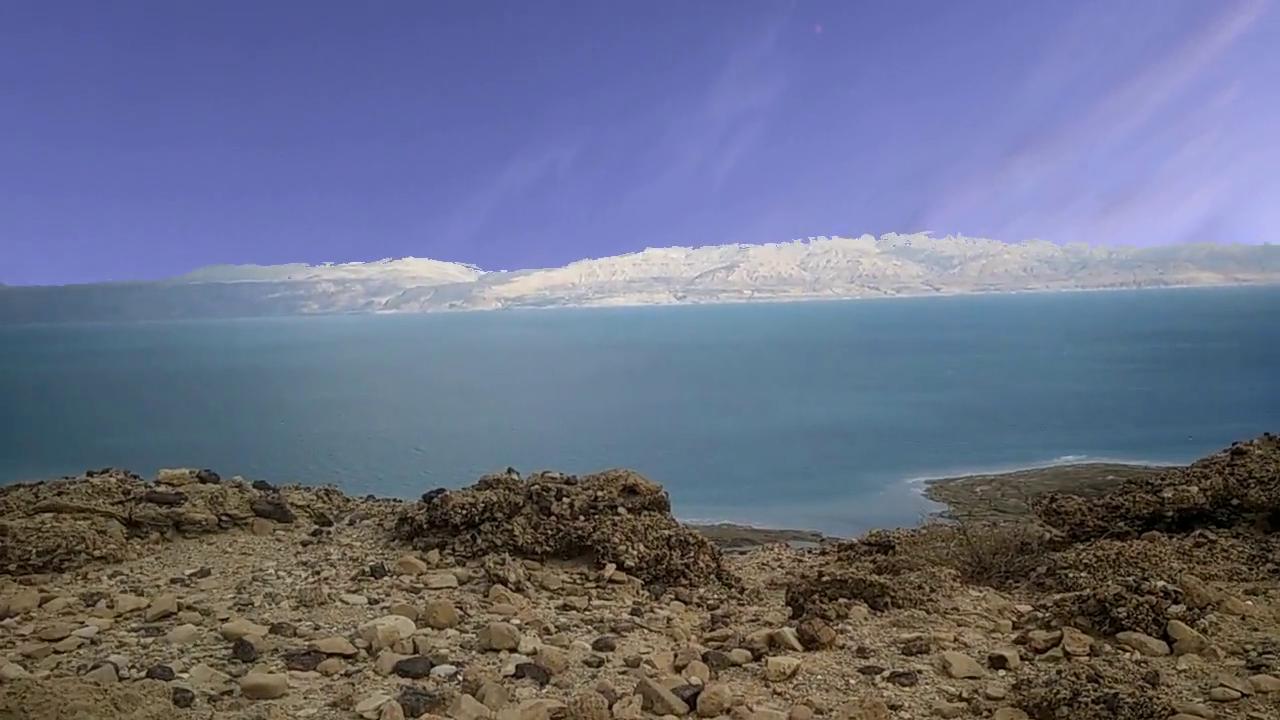}} &
{\includegraphics[width=0.5\linewidth]{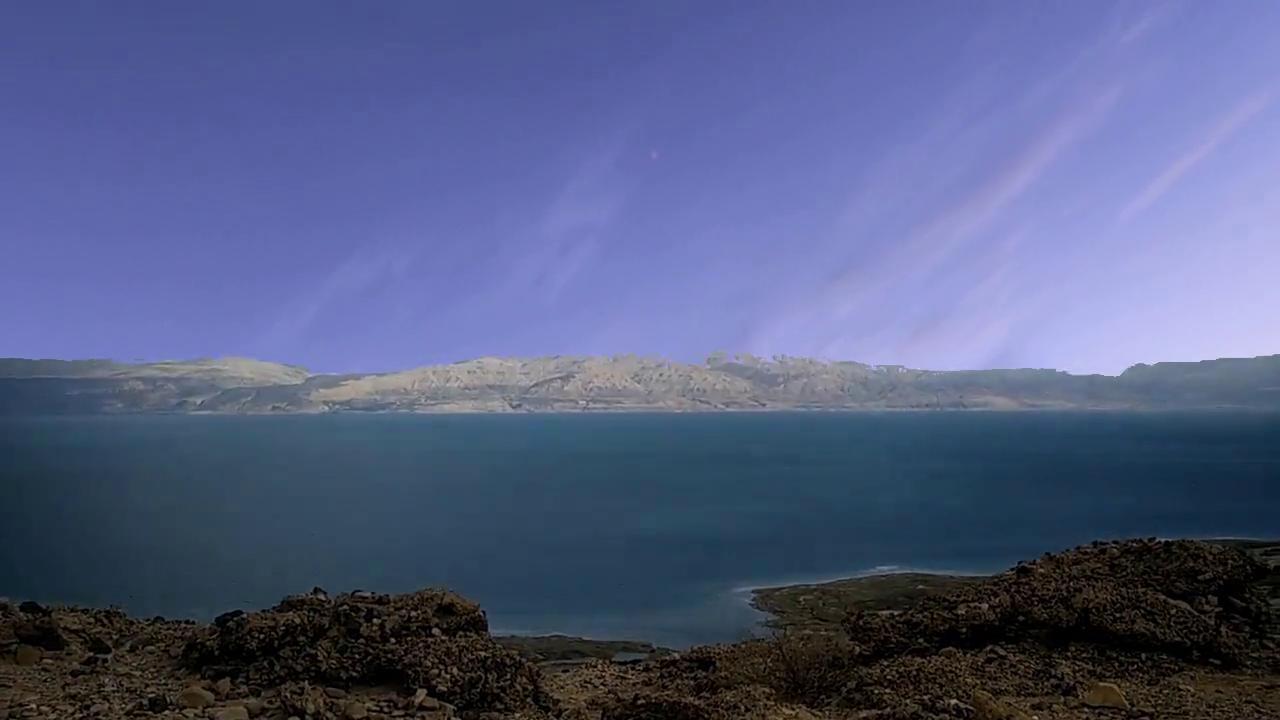}} \\
{\includegraphics[width=0.5\linewidth]{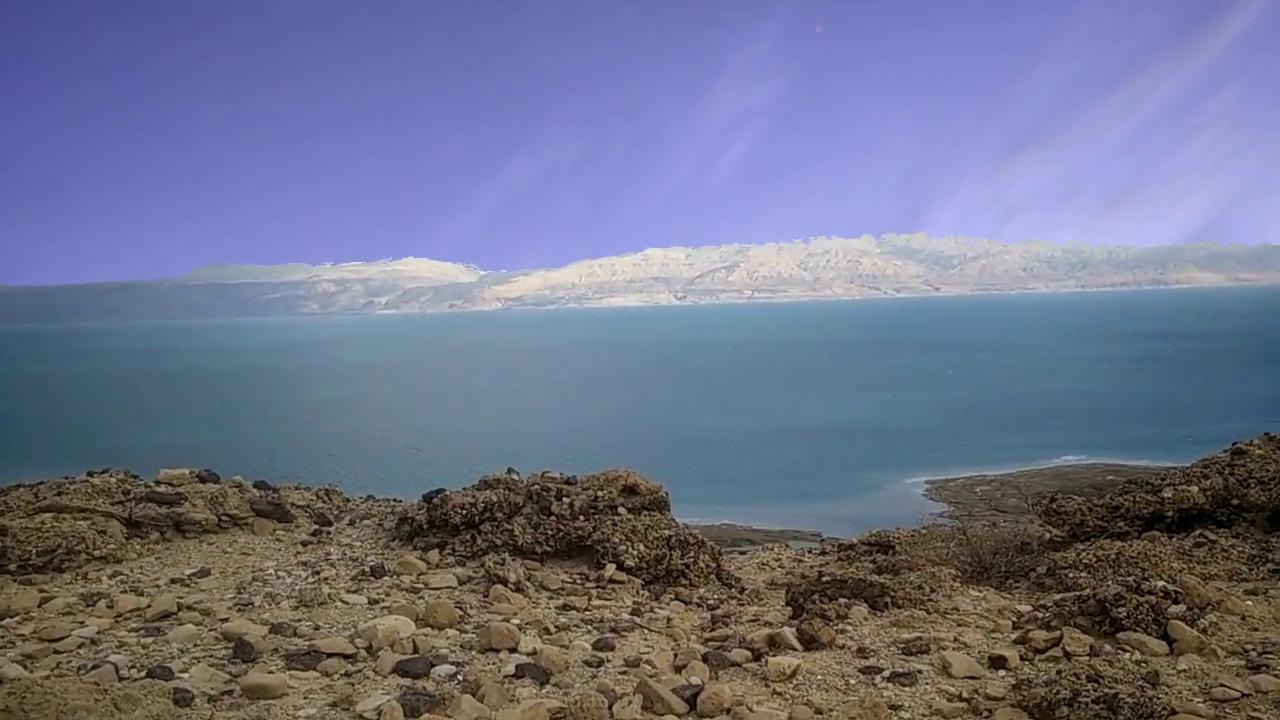}} &
{\includegraphics[width=0.5\linewidth]{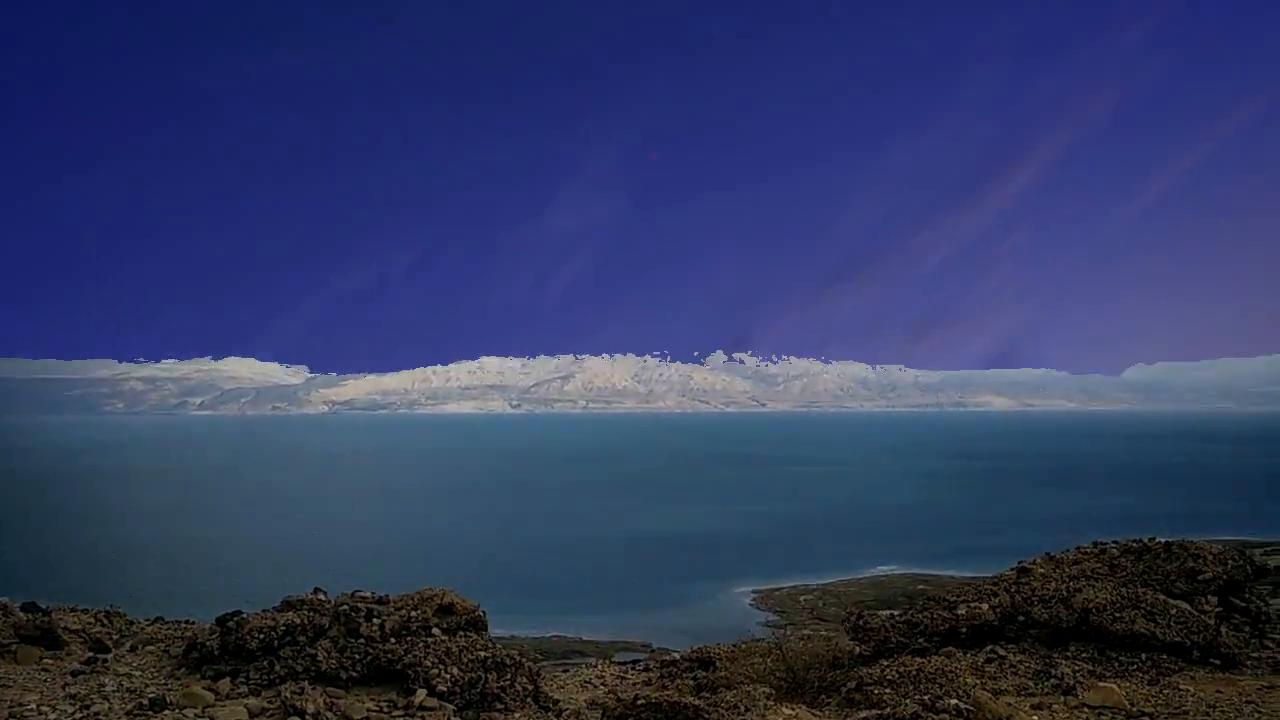}} \\
\end{tabular}
\caption{Vignetting and exposure. \textbf{Top}: two frames from original video captured with varying exposures. \textbf{Middle}: sky replaced with no exposure compensation. \textbf{Bottom}: exposure compensation and vignetting applied to the replaced sky.
}
\label{fig:vignetting}
\end{figure}

\section{Replacing the Sky}
A \coverVid video with the same camera motion and FOV as the \baseVid video is generated either from a spherical video or a still spherical image, often produced as a panoramic image. Perspective images can be reprojected from a spherical image with arbitrary FOV and pose, where the relation between different reprojected images is pure rotation. Thus, it is straightforward to generate a video with camera motion and FOV mimicking those of the \baseVid video. Interestingly, reprojecting from a single image looks quite natural as the viewer expects a static sky, at least in the short term. Moreover, skies are cropped from different parts of the \coverVid image to follow the motion of the \baseVid video (see different crops in Figure \ref{fig:pipeline}), together with applying exposure changes and vignetting (see Section \ref{sec:exposure}). As a consequence, sky video generated from a single image rarely has a 'frozen' feel.

An advantage of using a single \coverVid image, as opposed to a \coverVid video, is the reduced memory usage. Another is the relative paucity of available $360^{\circ}$ videos compared to that of still $360^{\circ}$ images. These advantages tend to be of even more importance when we address the resolution issue. Suppose we would like to replace the sky in a \baseVid video whose resolution is $1280x720$ pixels taken by a camera with horizontal FOV of $65^{\circ}$. To obtain a perspective reprojected sky from the \coverVid image with the same resolution, the horizontal size of the spherical \coverVid image should be at least $1280\frac{360}{65}=7089$ pixels. Fortunately, it is common for spherical images to be taken with such high resolutions. 

The \coverVid image may also be partial, for example a panoramic image create by combining images in various directions. As long as it covers all angles viewed by the \baseVid video it may be used for   sky replacement.

Even though the motion of the video reprojected from the \coverVid image is fully dictated by the motion of the \baseVid video, there are three degrees of freedom left in choosing the starting camera pose (pan, tilt, and roll). As there is no preferable starting panning direction,  it is left to the user.
The tilt and roll values  need to match those of the first frame of the \baseVid video. There are a number of works aiming to determine pitch and roll from a single image (see for example \cite{workman2016horizon}). We, however, did not incorporate such a method as its prediction can only be given to the user as an initial guess and must be refined anyways.

We merge an image $I_n$ from the \baseVid video with an image $J_n$ generated from the \coverVid image (if \coverVid video is used, both videos should have the same frame rate) by first reprojecting the latter from an equirectangular projection to a perspective camera projection with the FOV of $I_n$, according to the recovered camera rotation. The merge uses the segmentation mask $M_n$ as an alpha channel $OUT_n=(1-M_n)I_n+M_nJ_n$. 

To embed the \coverVid image's sky naturally in the \baseVid video we apply its exposure changes and camera vignetting to the transferred sky pixels. We then apply tonal adjustments to further harmonize the combined layers.

\subsection{Exposure Variation and Vignetting}
\label{sec:exposure}

Similarly to tracking, exposure and vignetting estimation is based on the luma channel. We adopt the model presented by \cite{goldman2005vignette}, 
\begin{equation}
P_i^j=g(e_iV(x)L^j)
\end{equation}
where $g$ is the Camera Response Curve (CRC), $e_i$  the relative exposure of frame $i$, $V(x)$ is vignetting per spatial location, and $P_i^j$ is pixel intensity indexed by frame $i$ and imaged object $j$ whose radiance is $L^j$. We use the cost function presented in \cite{bergmann2018},
\begin{equation}
E=\sum\limits_i\sum\limits_j w_i^j\|P_i^j-g(e_iV(x)L^j)\|_h
\end{equation}
where $w_i^j$ depend on edge intensity and $h$ is the Huber norm parameter. The under-constrained problem of simultaneously estimating vignetting, CRC, exposure and radiance is solved by  coordinate descent on the four parameters. Similarly to \cite{bergmann2018} we minimize this function using Levenberg-Marquardt with an analytic Jacobian, except for CRC's Jacobian which is calculated numerically as it is non parametric and learned from data (we used the values provided by \cite{grossberg2004modeling}). 

Generally, without vignetting estimation the problem is badly conditioned, as CRC recovery  depends highly on the existence of strong changes in exposure \cite{kim2007joint}. Thus, we minimize the cost function over pixels from a subsequence with intense camera motion in which the vignetting effect is substantial and the optimization enables reliable recovery of CRC. To estimate vignetting, correspondences over large spatial range are necessary. Therefore we subsample tracked trajectories which are both long and scattered over the entire image frame. This is also beneficial for the estimation of the CRC as pixels across the entire intensity curve  participate in the estimation. 

Since vignetting and CRC do not change during the video, once they are computed for a small subset of frames we fix them and optimize for exposure change through the rest of the video by minimizing $w_i^j\|e_iL^j-g^{-1}(P_i^j)/V(x)\|$ over exposure and radiance. Linear optimization is performed by batch coordinate descent  fixing either exposures or radiance and calculating the other. Usually it only takes a few iterations to converge. Non linear Levenberg-Marquardt optimization for this second stage yielded only marginal improvement. 

To transfer vignetting and exposure changes from the \baseVid video to the reprojected \coverVid video, after cropping an image $J_n$ from the \coverVid video, we apply the exposure $e_n$ computed from $I_n$ and vignetting to every pixel $P$ using the estimated CRC of the \baseVid video 
\begin{equation}
\overline{P}=g(g^{-1}(P)e_nV(x))
\end{equation}
The superscript is omitted to point out that the radiance of this pixel has no effect on the intensity transfer. Ideally, we would use the inverse CRC of the \coverVid image. However, it is usually unknown, as we only have a single image. Instead, we apply the CRC of the \baseVid video to the projected sky to make its changes in accordance with the changes of the \baseVid video.
See Figure \ref{fig:vignetting} for an example of exposure changes applied to the replaced sky.

\begin{figure}
{\includegraphics[width=1\linewidth]{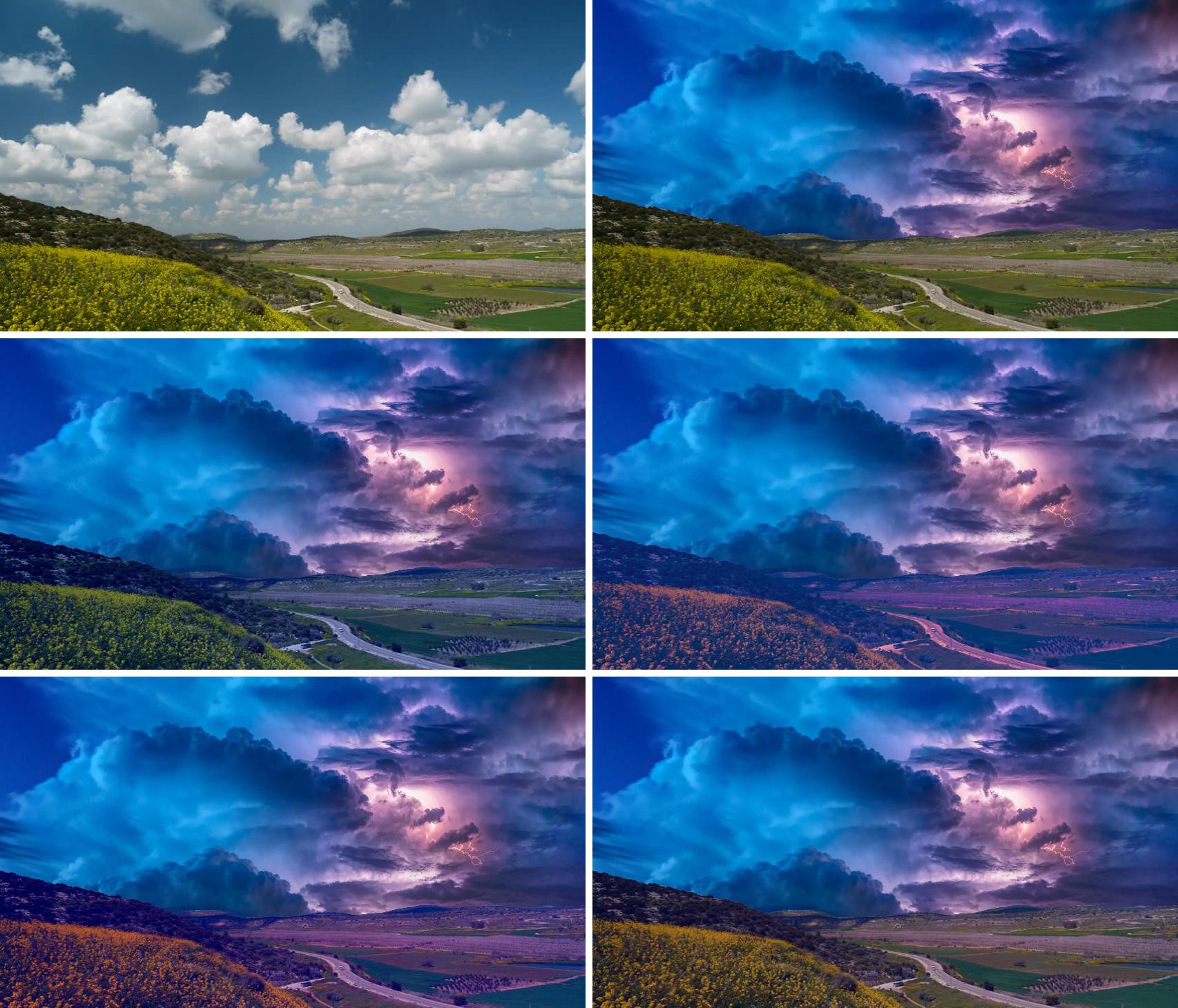}} 
\caption{Comparing color transfer methods. \textbf{Top:} Original and simple composite images. \textbf{Middle:} Color transfer using \cite{reinhard2001color} (left) and \cite{Piti2005NdimensionalPD} (right). \textbf{Bottom:} Color transfer using MKL \cite{Piti2007TheLM}  (left) and our final result after blending MKL with the original image (right). Blending the original foreground with one that has a color histogram resembling that of the new skies, achieves more realistic results by simulating the new airlight component of each pixel in the foreground.}
\label{fig:color_transfer}
\end{figure}

\subsection{Color Transfer}
To look realistic, the lighting, color histogram and other tonal properties of the \baseVid video and the \coverVid video need to be aligned. We use tonal manipulations in both directions in order to change the airlight in the \baseVid video from the original airlight to the one created by the newly replaced sky as well as propagate haze effects from the \baseVid video onto the \coverVid video. To adjust the global airlight in the \baseVid video we use the affine Monge-Kantorovitch color histogram transfer algorithm \cite{Piti2007TheLM} to transfer the histogram of the \coverVid from the sky region only (using the segmentation) onto the histogram of the \baseVid video in its entirety (including the soon to be replaced sky region). To allow for less computations as well as temporal consistency, the Monge-Kantorovitch matrix is re-calculated every 8 frames and  interpolated between them. While the Monge-Kantorovitch color transfer is not as exact (in the sense of reproducing the color histogram of the reference image) as for example the sliced Wasserstein method of \cite{Piti2005NdimensionalPD}, it is much faster. The resulting composited videos often have a much more natural look with a less bimodal color histogram (see Fig. \ref{fig:color_transfer}). To mitigate the effects of haze, which appear not only in the sky region but also in the rest of the image and most prominently near the horizon (e.g. Fig. \ref{fig:results2} (b)), we estimate the horizon line in the \baseVid video via the segmentation mask and propagate the lightness from that region into the sky region in the \coverVid video.

\begin{figure}
\begin{tabular}{cc}
{\includegraphics[width=.45\linewidth]{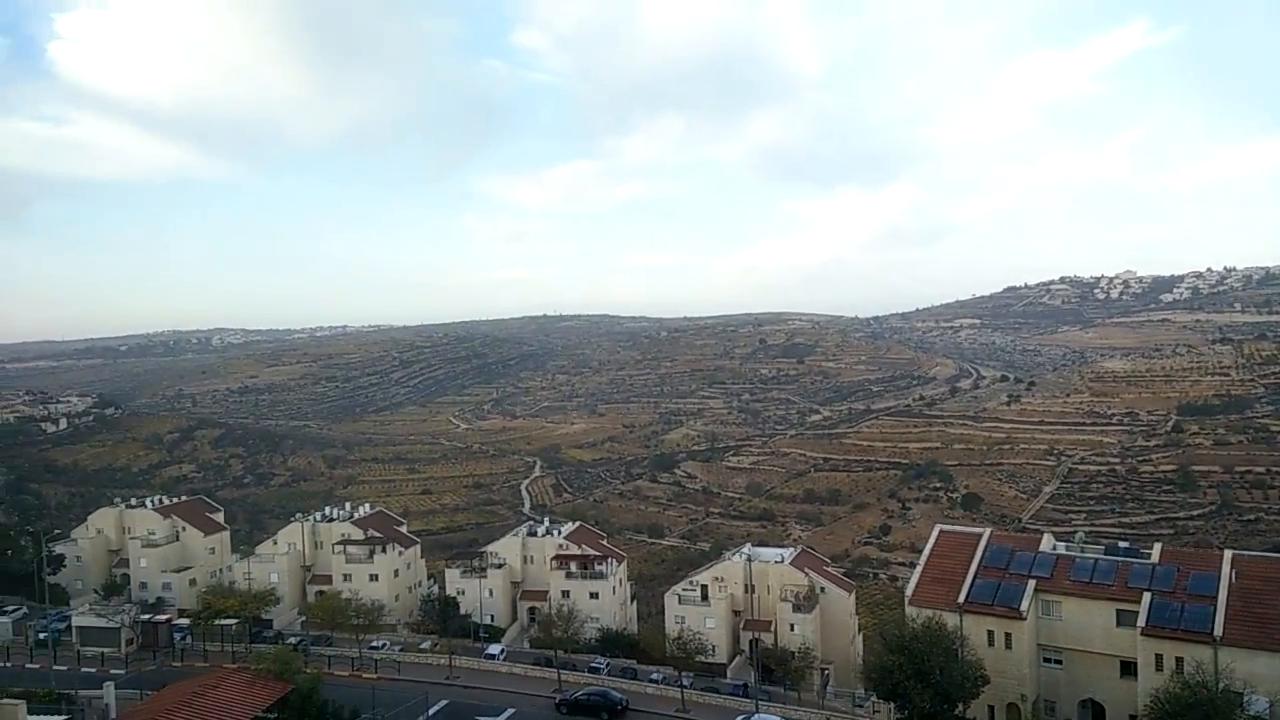}} &
{\includegraphics[width=.45\linewidth]{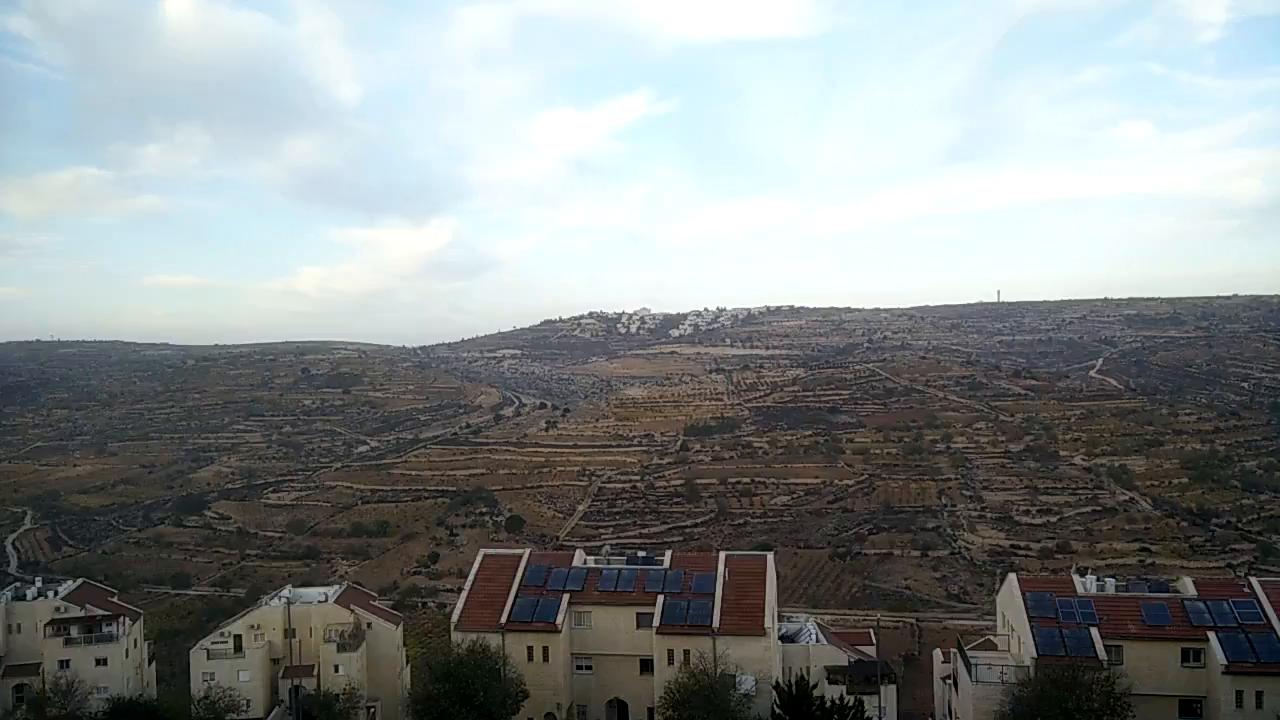}} \\
\multicolumn{2}{c}{
{\includegraphics[width=0.9\linewidth]{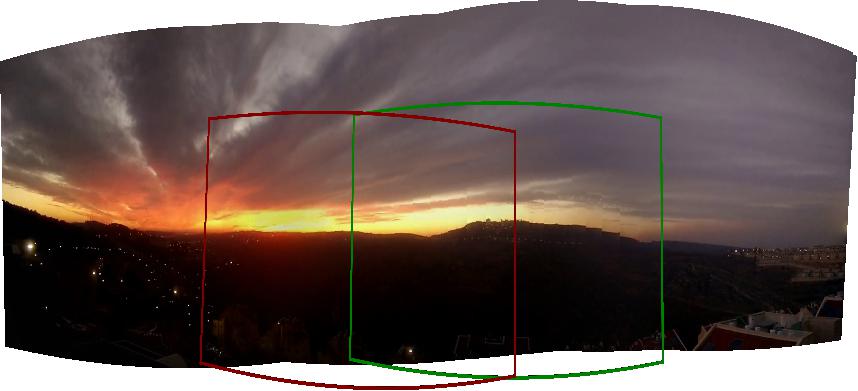}}} \\
{\includegraphics[width=.45\linewidth]{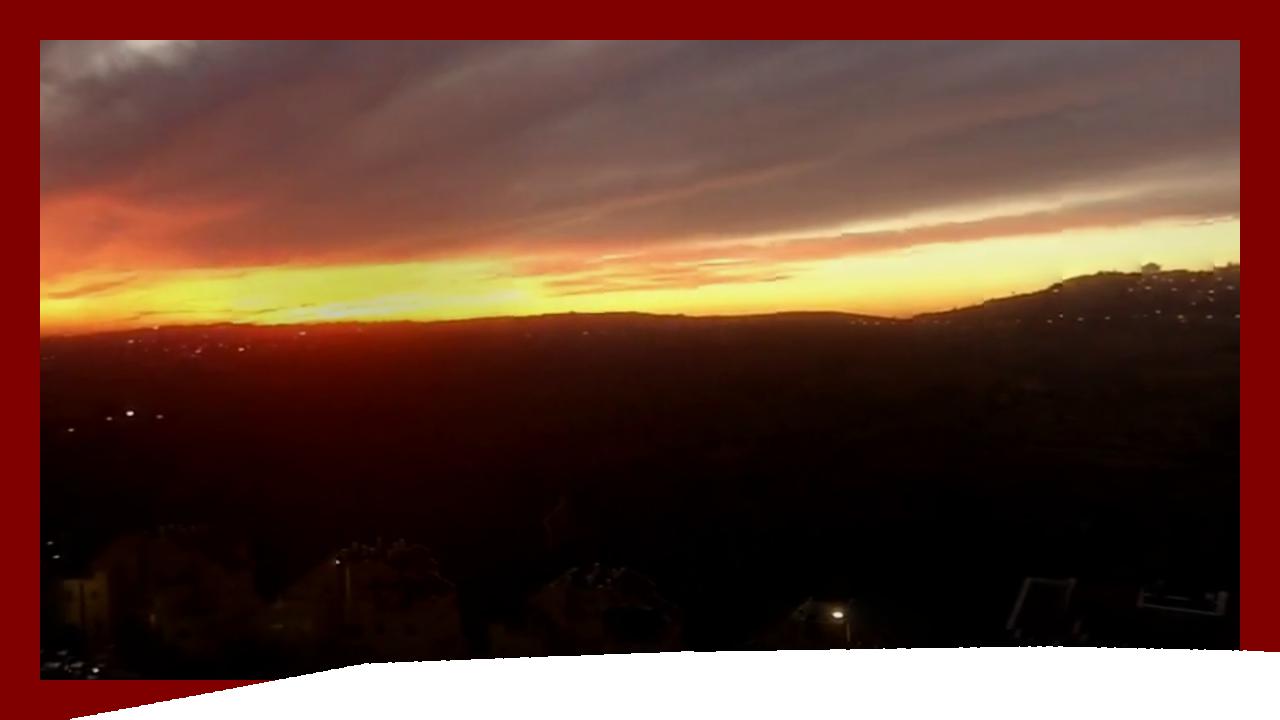}} &
{\includegraphics[width=.45\linewidth]{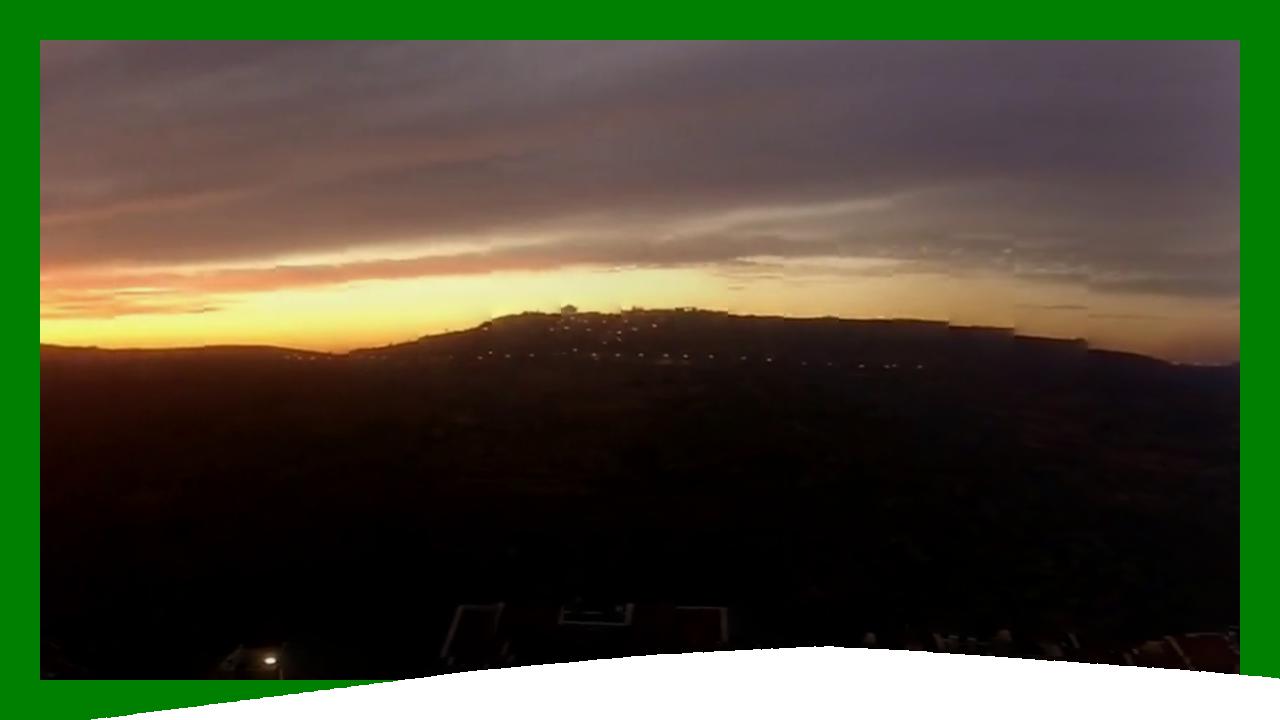}} \\
{\includegraphics[width=.45\linewidth]{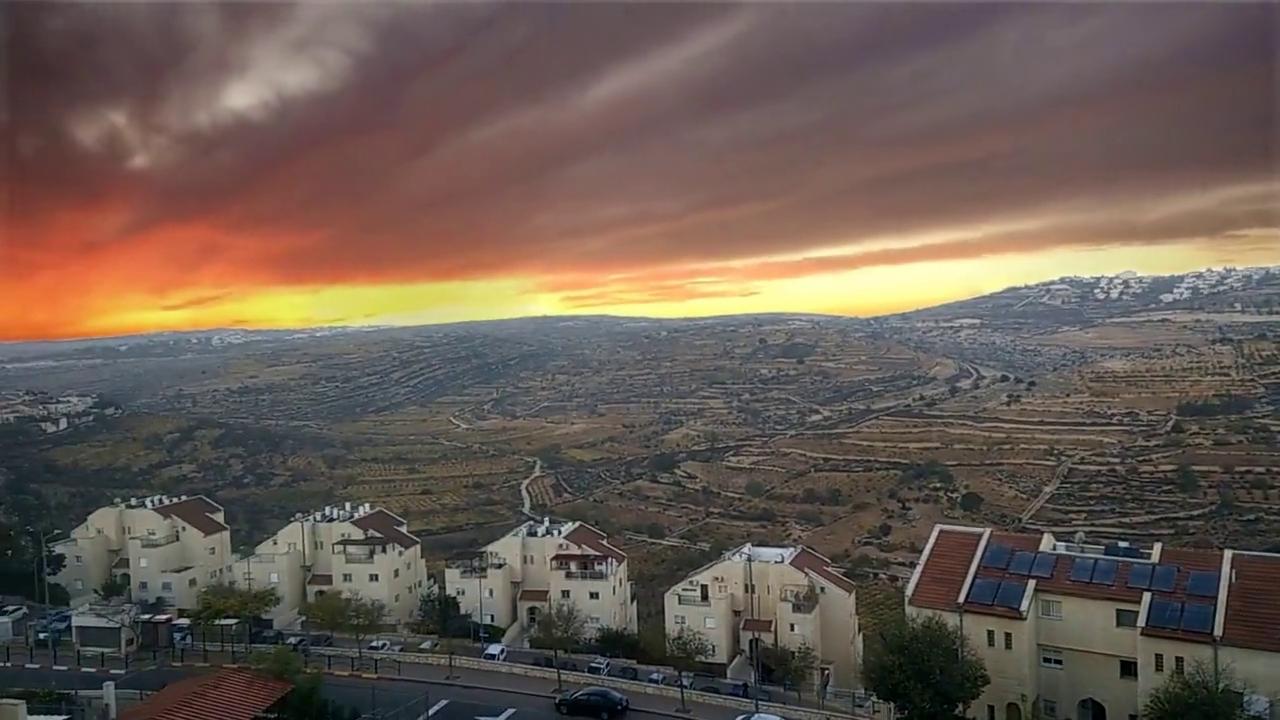}} &
{\includegraphics[width=.45\linewidth]{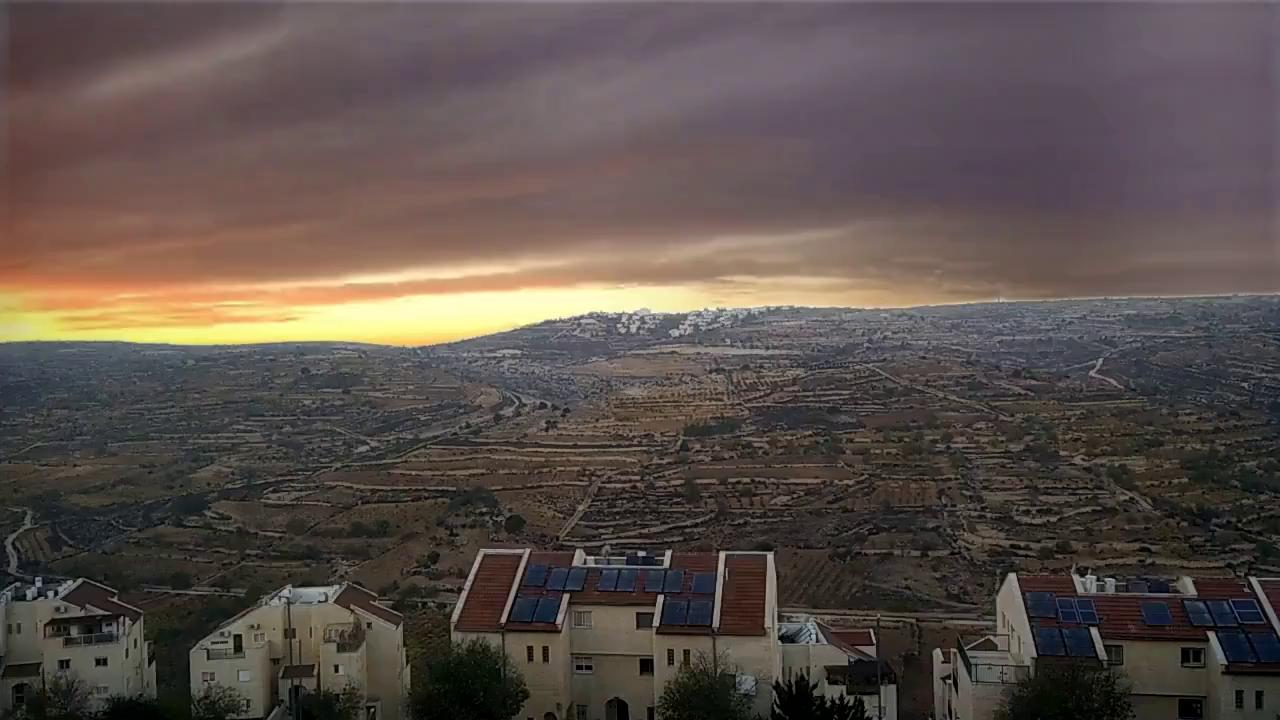}} \\
\end{tabular}
\caption{Video HDR. \textbf{Top row:} two frames from \baseVid video. \textbf{Second row:} Spherical image created from another video captured at the same location during sunset and used as the \coverVid image. \textbf{Third row:} new perspective images sampled from the \coverVid image according to the motion in the \baseVid video. \textbf{Bottom row:} frames from output HDR video. Please refer to supplementary video for full sequences.}
\label{fig:hdr}
\end{figure}

\section{Results and Analysis}
\label{sec:results}
Please see sample frames of algorithm results in Figures \ref{fig:results}, \ref{fig:results2}. Additional video clips are provided in the supplementary material \footnote{https://youtu.be/1uZ46YzX-pI}.

\subsection{Network Training Details}
We used the Adam optimizer \cite{Kingma2014AdamAM} with exponential learning rate decay. We trained our network for 50 epochs, based on observed convergence rates for this task.

We compared ourselves to \cite{Tsai2016SkyIN}, using 1045 random images exhibiting the sky or the cloud sky classes out of the LMSun \cite{Tighe2010SuperParsingSN} dataset, similar to the evaluation process in \cite{Tsai2016SkyIN}. 

The model used to produce the IOU statistics reported below does not have a fourth feedback input channel, as this testing was done on an image dataset and not videos. It had 8 residual blocks with a residual bottleneck of 32 filters. 

When calculating the average mean-IOU ratio on these images between the binarized raw network output (binarized with respect to a threshold of 0.5) and the ground truth, we report an average IOU of 88.8\%, higher than the 87.6\% reported before refinement in \cite{Tsai2016SkyIN}. Moreover, 69.0\% of the images achieve an IOU ratio higher than 90\%, which is considered visually pleasing, a considerably higher ratio, even without refinement, than the approximately 62\% reported after refinement (as estimated from Figure 5 in \cite{Tsai2016SkyIN}). To evaluate the accuracy and temporal consistency of our feedback model we conducted the following experiments: $(i)$ We generated videos from images with ground truth sky segmentation using a virtual camera path; $(ii)$ We measured IOU and compared it to the feedback model, and to the same model without a dense conditional random field (CRF)\cite{Krhenbhl2011EfficientII}, using  the CRF code provided by the authors. $(iii)$ We measure and report temporal consistency by projecting pairs of consecutive segmentation masks onto the same plane (we have the ground truth transformations) and measuring their difference.

The accuracy does not alter significantly by adding the feedback channel. After applying CRF to refine the results, the resulting binary maps achieve an average IOU score of 88.1\% with respect to the ground truth masks, comparable with the 88.7\% average IOU reported in \cite{Tsai2016SkyIN} (though they use their own refinement procedure). 67.3\% of the images achieve an IOU ratio of 90\% or higher, considerably more than in \cite{Tsai2016SkyIN}. Although numerically the CRF slightly degrades accuracy as measured against the ground truth segmentation, subjectively the results look better. This counter intuitive observation can be explained by the ground truth annotations of sky regions with difficult boundaries, such as trees, being very inexact, as it is composed of simple polygons roughly sketched by human annotators. 

Temporal consistency improved significantly  when adding a feedback channel,  by a factor of $2.06$, and by a factor of $2.41$ after applying a CRF. Yet, in some scenes temporal consistency artifacts are visible, especially in highly complex scenes or due to a strong lighting change when the camera rotates (e.g. towards the sun).

\subsection{Residual ablation study}
\begin{figure}
{\includegraphics[width=1\linewidth]{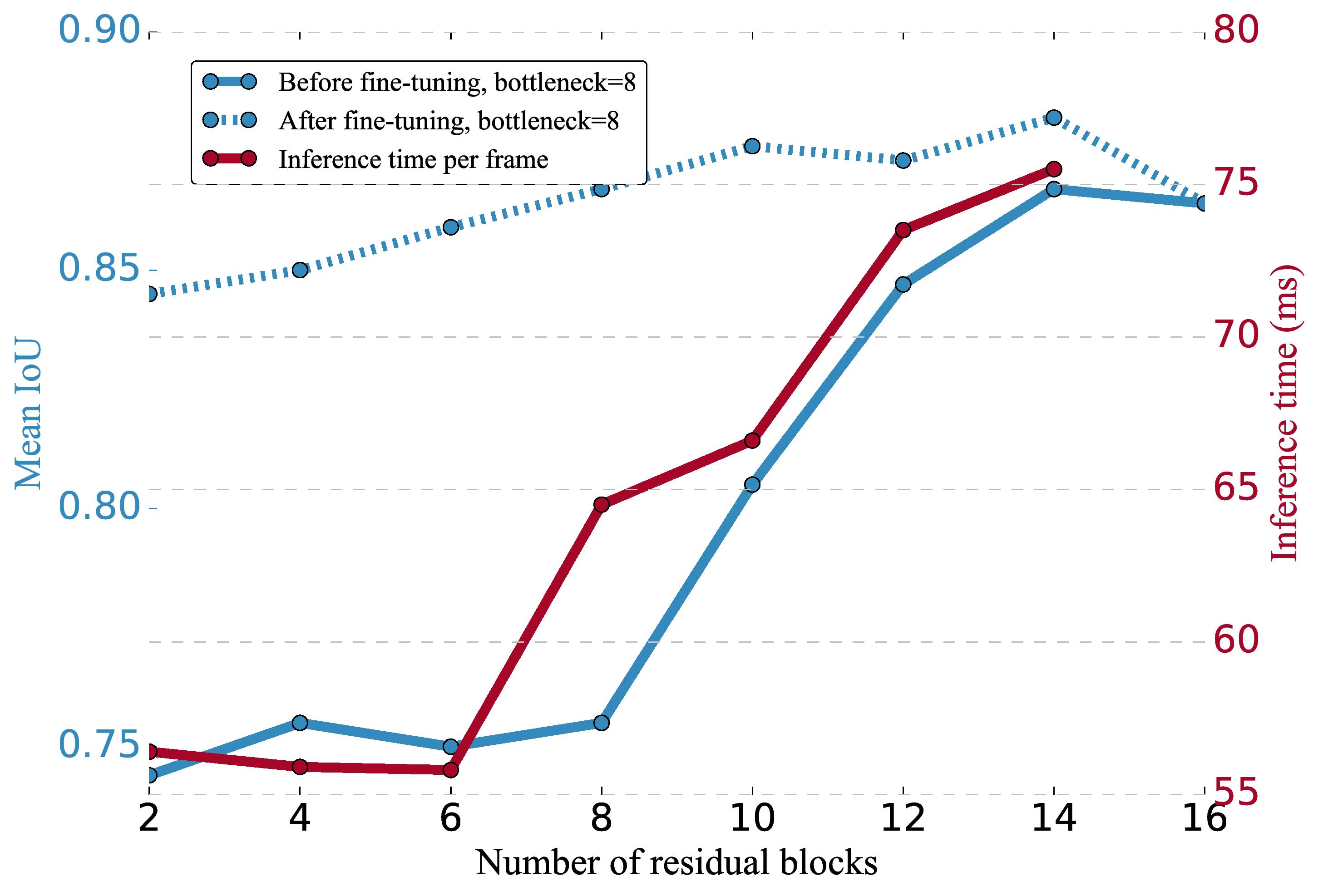}} 
\caption{Effects of removing residual blocks on segmentation accuracy and running time. Reducing the number of blocks decreases running time and accuracy in correlation. However, after fine-tuning accuracy climbs almost to the same level as the full model, suggesting that in our model a good trade-off is achieved with 6 residual blocks.}
\label{fig:residual_abelation}
\end{figure}

To study the effect of adding more residual blocks or dropping some of the last ones to speed up the inference stage, we trained three models, differing in their number of residual blocks (8, 12, 16) and in the size of their residual bottleneck (32, 16, 8 respectively) on the same data with the same optimizer and the same number of epochs.  
We then calculated the mean IOU value on the 1045 LMSun images for  ablated versions of these models where the last residual blocks hae been skipped. As is evident in Figure \ref{fig:residual_abelation}. As expected, the average IOU measured increases monotonously with the number of residual blocks, albeit most of the improvement is demonstrated by specific residual blocks.  

We then also fine tuned these ablated versions for a fixed number of additional epochs and with a reduced learning rate, and again measured the resulting IOU. Again, an increasing monotonous relation exists but now its slope is smaller, as the additional fine tuning improved the performance of the ablated models. 

Finally, by considering the relative running times of the ablated models, one can then pick a desired trade-off between performance, which is especially important in video, and model accuracy.

\begin{figure}
{\includegraphics[width=1\linewidth]{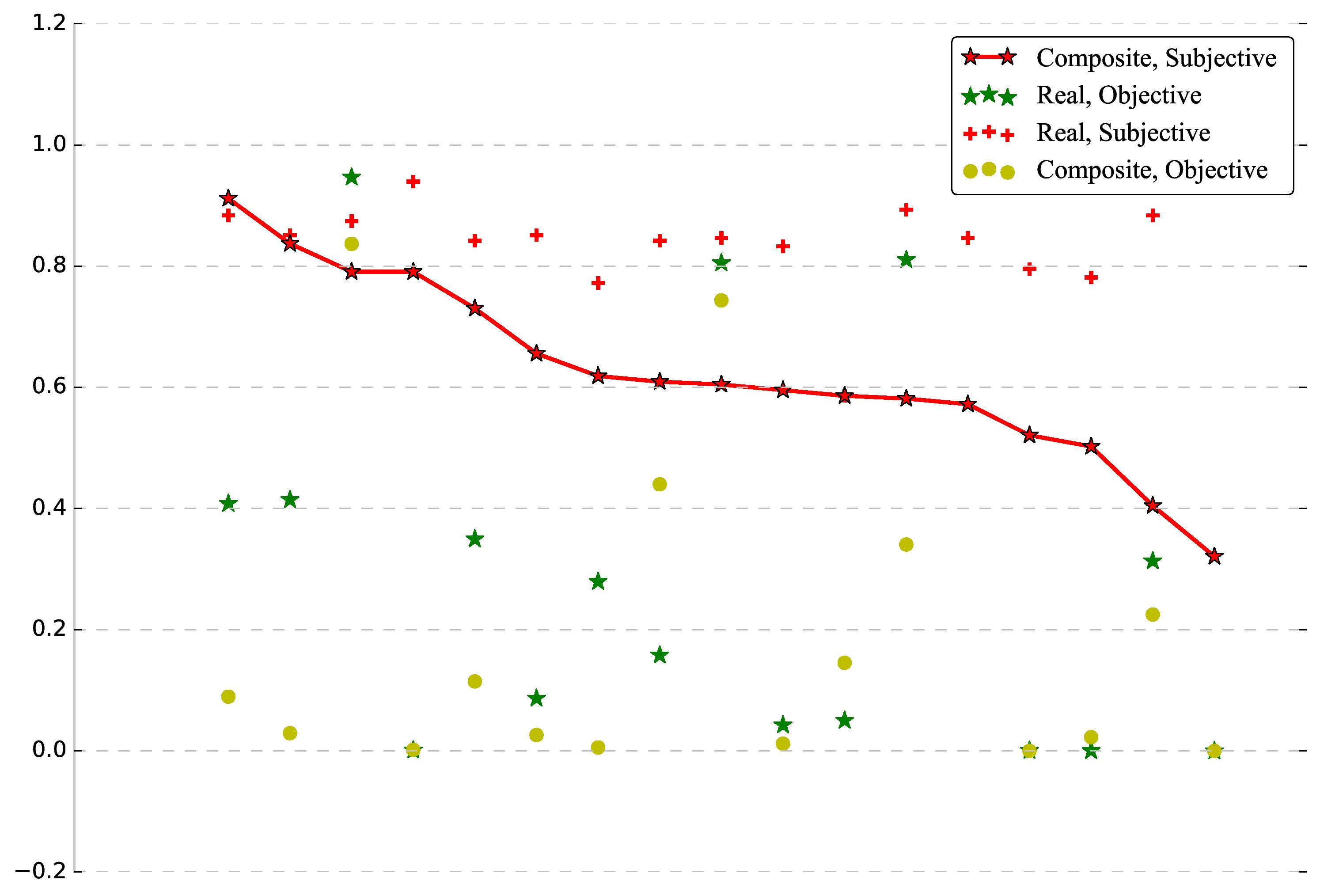}} 
\caption{Quantitative and qualitative evaluation of our method. We evaluated the realism of 17 videos before and after sky replacement, both based on a user study and by automatic means. Videos are ordered in decreasing score obtained from user study on replaced videos. Most of the replaced videos got realism score >= .6 and were not far behind original videos based on this measure. Subjective scores were normalized from the range [1,5] to [0,1].}
\label{fig:scores}
\end{figure}

\begin{figure*}
\begingroup
\begin{tabular}{cc}

(a) & {\includegraphics[width=0.9\linewidth]{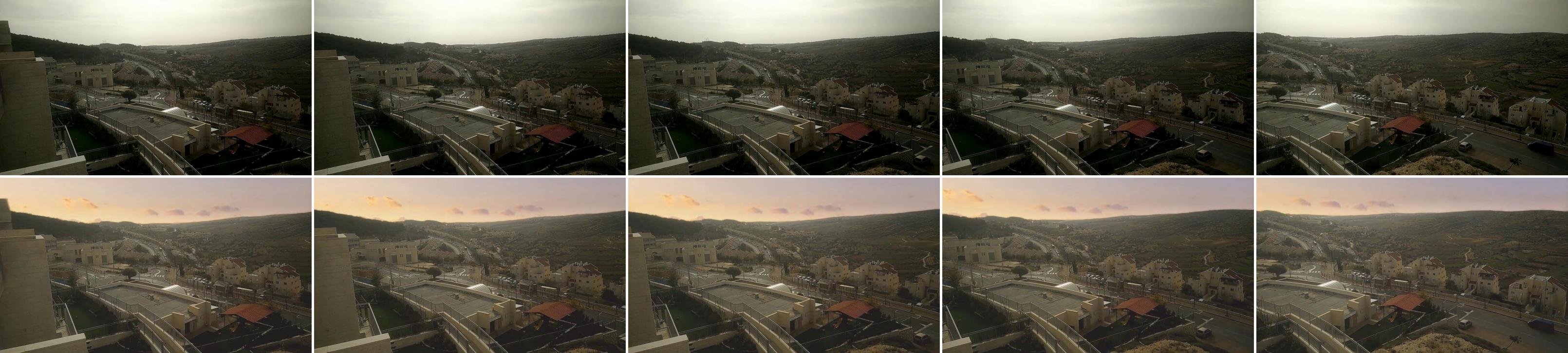}} \\
(b) & {\includegraphics[width=0.9\linewidth]{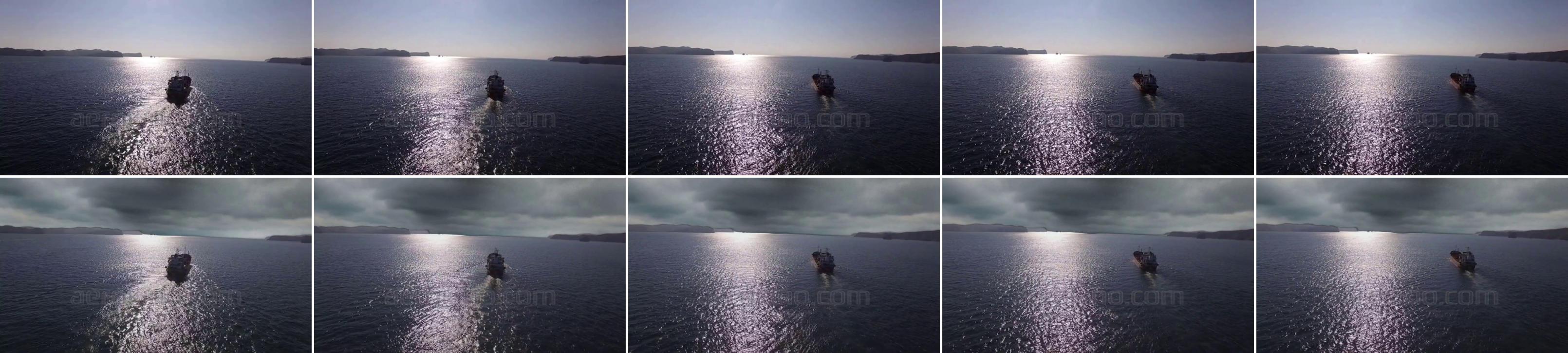}} \\
(c) & {\includegraphics[width=0.9\linewidth]{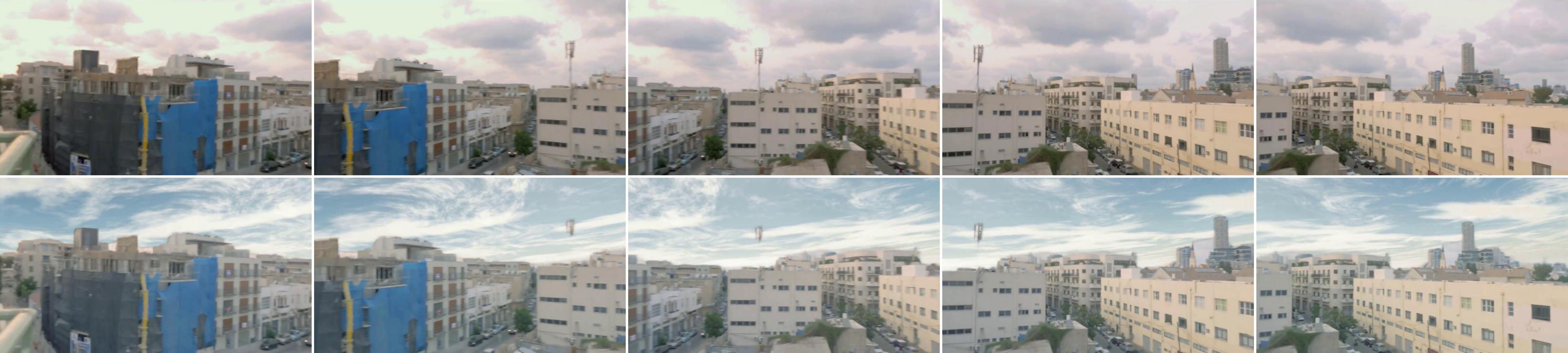}} \\

\end{tabular}
\caption{Sample frames from videos with replaced sky. Row pairs show corresponding frames from before and after sky replacement.
}
\label{fig:results}
\endgroup
\end{figure*}

\subsection{User study}
To assess the perceptual quality of our results we conducted a user study on a set of 17 real videos and the same videos after sky replacement. Every one of our 43 participants was asked to rank the realism of those 34 video clips, one at a time in random order. The participants were asked to assess the perceptual quality and realism of each clip independently on a scale between 1 and 5. The average score of real videos was $4.1$, and of our composite videos $3.1$. 
Scores for individual videos are shown in Figure \ref{fig:scores}.

\subsection{Automatic realism score}
In addition to the subjective evaluation we performed a more objective one. For this, we used a CNN trained to distinguish real photographs from composite images \cite{zhu2015learning} which was shown to correlate highly with human perception. This RealismCNN outputs a realism score in the range $[0,1]$ for a given image. For videos, we computed this score per frame and assigned the average as the realism score for the video. It is interesting to compare a score of a video with the score of its sky replaced version. On average real videos obtained a score of 0.32, while our composite videos were not far behind with 0.23.
Individual scores are provided in Fig. \ref{fig:scores}. 

\subsection {Running Time on Mobile Device}
We tested the algorithm's running times on an iPhone 6S. The chosen segmentation network takes 55 ms per frame using the GPU, evaluating all layers for every frame. Color transfer takes 10 ms.
KLT Tracking takes 10 ms. with OpenCV  \cite{opencv_library} and camera motion estimation takes about 2 ms. Substituting tracking with rotation measurements from the device's gyroscope (or sensor fusion \cite{you2001fusion}) will potentially provide faster and image independent rotation estimations. This implementation makes augmented reality applications such as live sky replacement while in video chat possible at a frame rate of 15 fps. If the \baseVid video is captured on the device its FOV may be provided by the device and the FOV calculation step is skipped.

\subsection{Video HDR}
Usually the sky is much brighter than the rest of the image. Our work naturally extends to creating a high dynamic range video. For this  we capture a couple of videos with different exposures, whose fields of view overlap on the sky region. One of them - typically the one with lower exposure, in which the sky is correctly exposed - is used to construct a spherical panoramic image using the method of  \cite{szeliski1997creating}, while the other serves as the \baseVid video, preserving scene dynamics. An example is shown in Figure \ref{fig:hdr}. We estimate vignetting, CRC and focal length on the \baseVid video, as the features are easier to track and apply the same values to both videos. 

A video which concentrates on the sky may be hard to track, due to the scarcity of  features in sky regions, and the fact that landscape regions may be underexposed. One might prefer direct iterative alignment \cite{baker2004lucas}. However, in our experiments, on some frames it did not converge, so we dropped this attempt. 

\begin{figure}
\begin{tabular}{cc}
{\includegraphics[width=.45\linewidth]{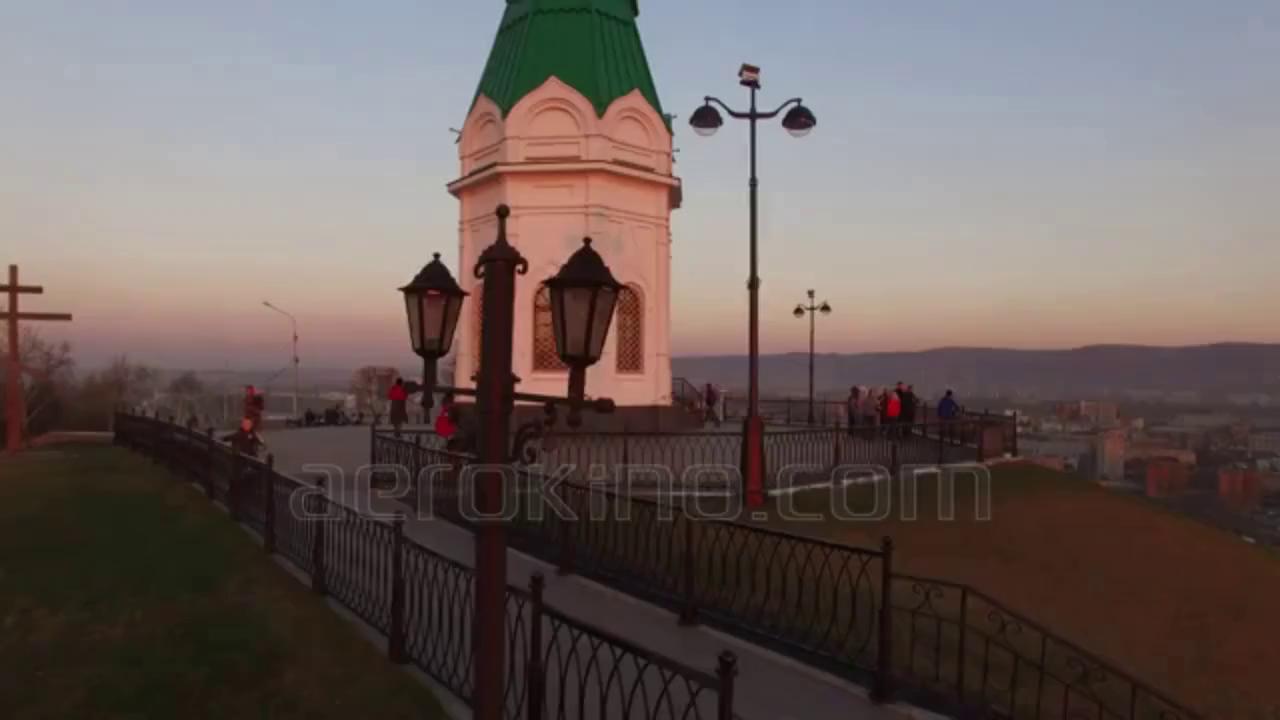}} &
{\includegraphics[width=.45\linewidth]{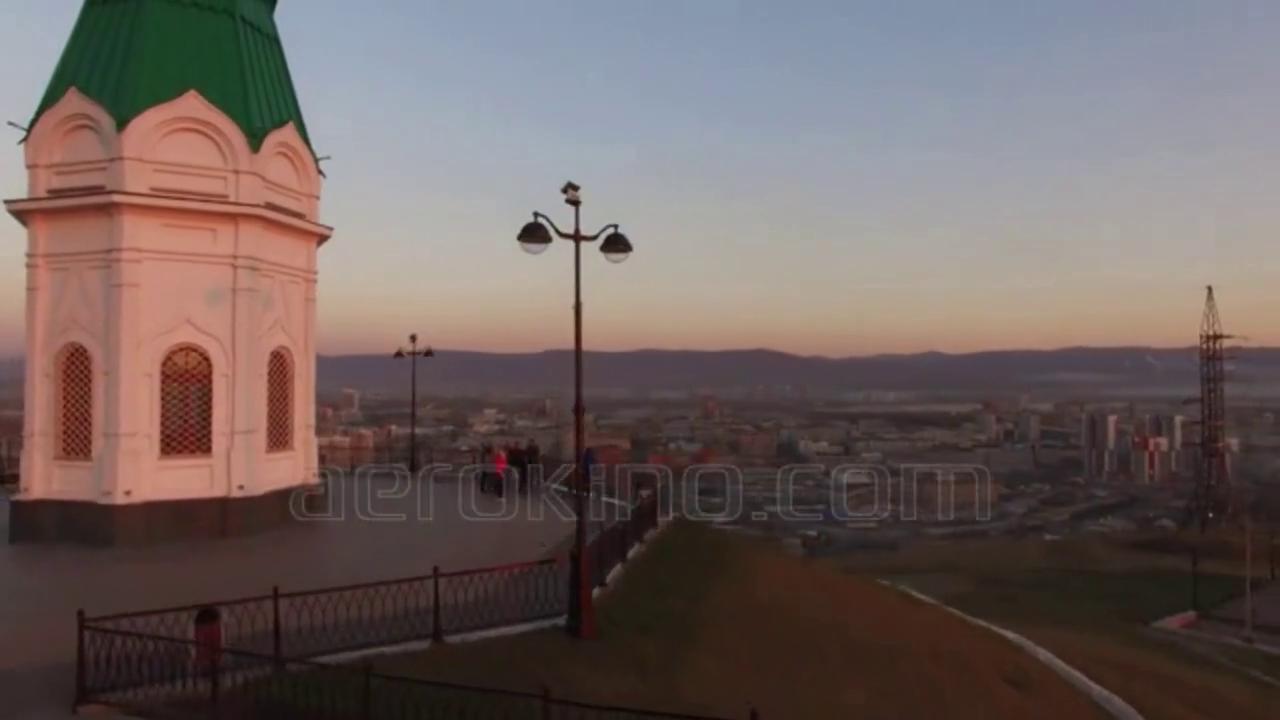}} \\
{\includegraphics[width=.45\linewidth]{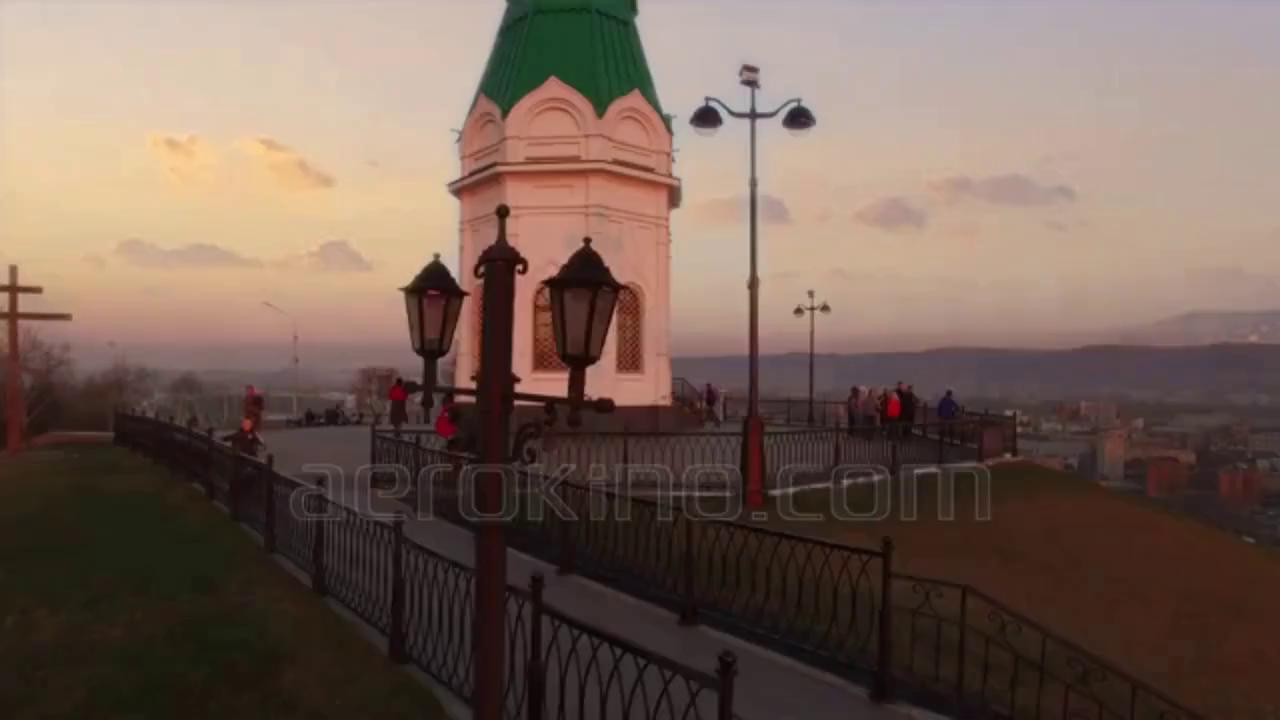}} &
{\includegraphics[width=.45\linewidth]{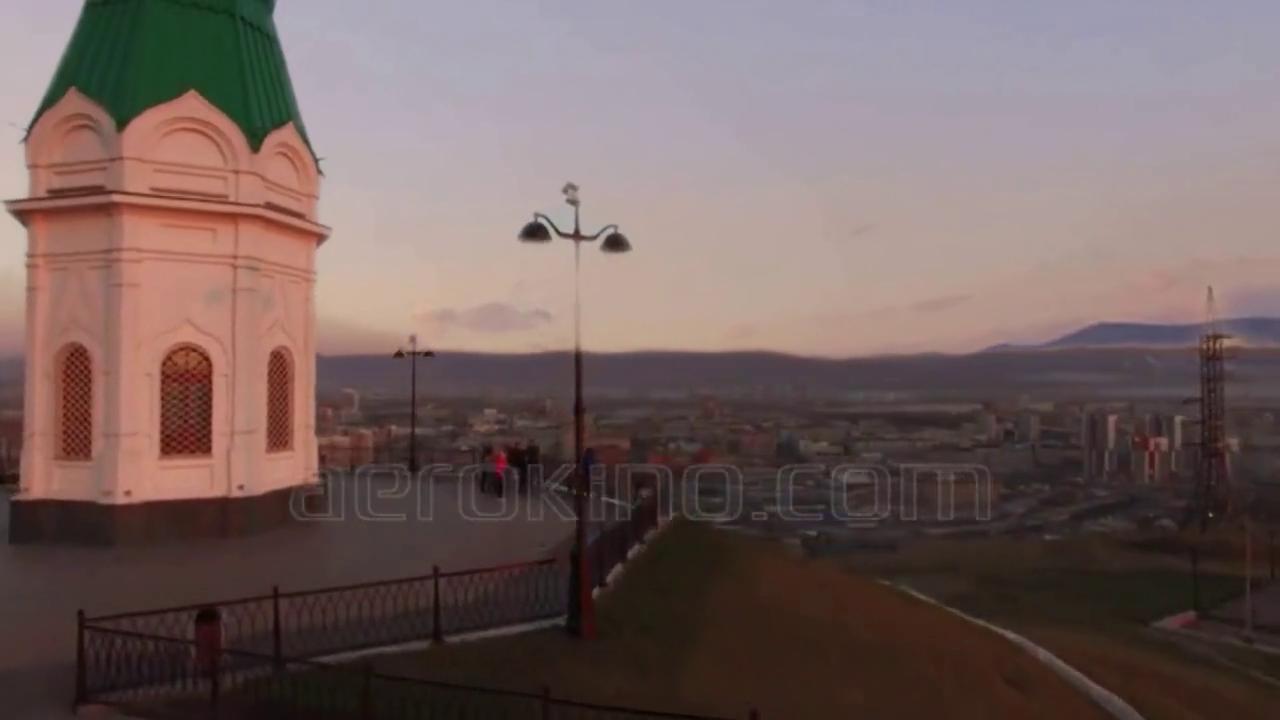}} \\
\end{tabular}
\caption{A substantial camera translation may result in inaccurate rotation estimation. \textbf{Top:} Images captured by a drone, where the majority of the motion is translational. \textbf{Bottom:}  Rotation of the replaced sky  differs from that of the original video.}
\label{fig:wrong_motion}
\end{figure}

\subsection{Limitations}
The assumptions that our design choices rely on do not always hold. The sky segmentation network, while producing consistent segmentation masks, might be consistent on errors as well. In some cases, a wrong segmentation region in the order of a few pixels in a single frame started to grow in consecutive frames through the feedback loop of the network. Also, isolated small foreground elements are sometimes segmented as sky, e.g the pole on the roof in Fig. \ref{fig:results} (c).

Another failure  case  is inaccurate camera motion estimation. Motion and FOV estimation assume a relatively small amount of translation. Under significant camera displacement, such as in videos taken by a drone, wrong motion estimation may lead to motion inconsistency between the replaced sky and the original video. An example for this type of  motion discrepancy is illustrated in Figure \ref{fig:wrong_motion}.

\begin{figure*}
\begingroup
\begin{tabular}{cc}
(a) & {\includegraphics[width=0.9\linewidth]{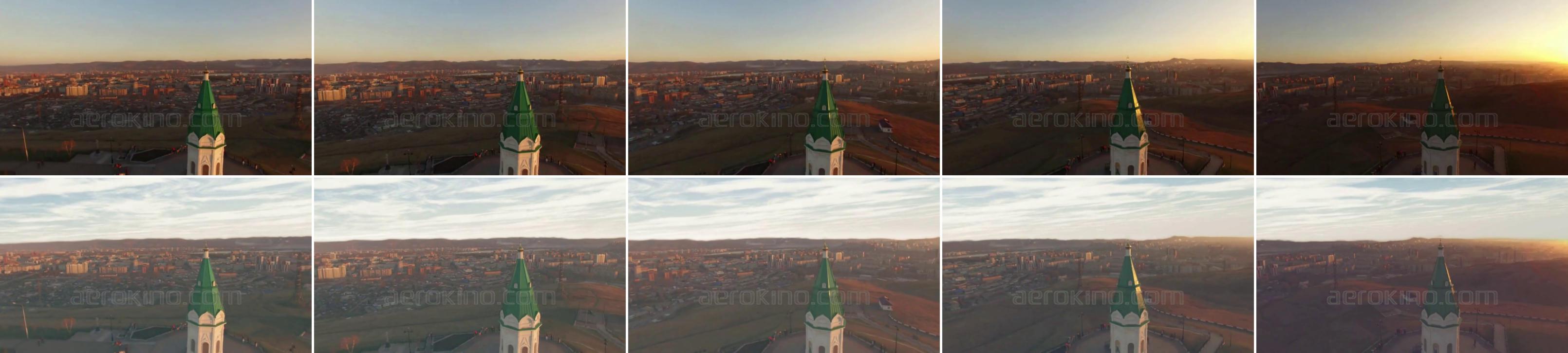}} \\ \\
(b) & {\includegraphics[width=0.9\linewidth]{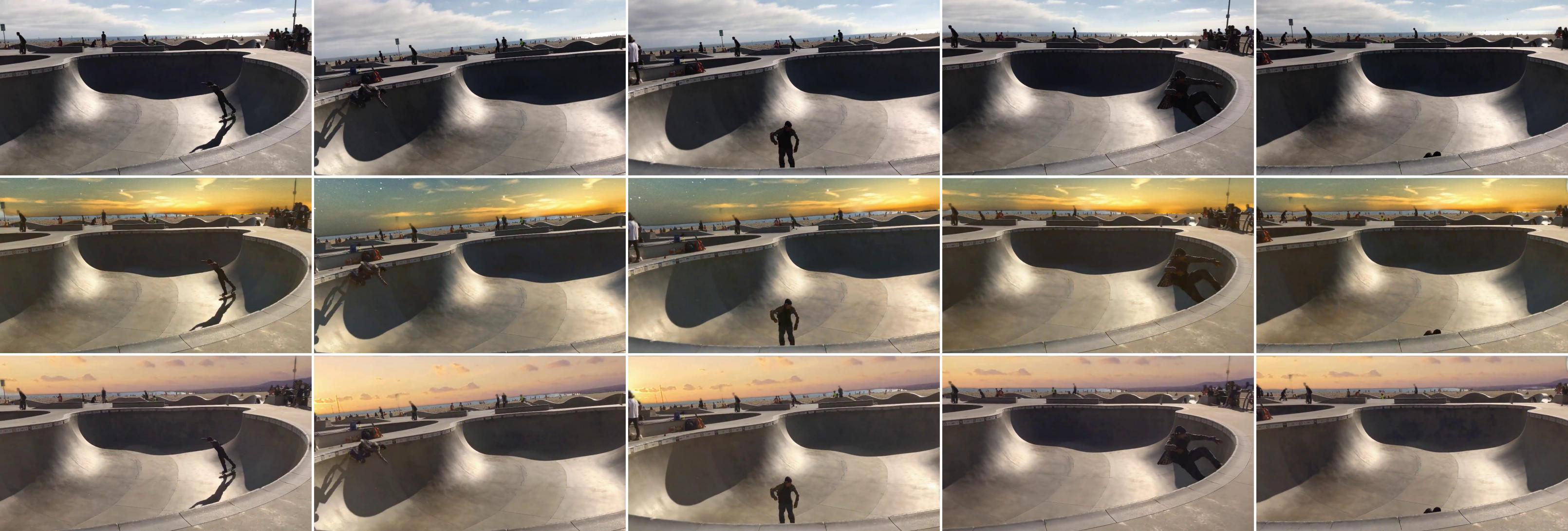}} \\ \\
(c) & {\includegraphics[width=0.9\linewidth]{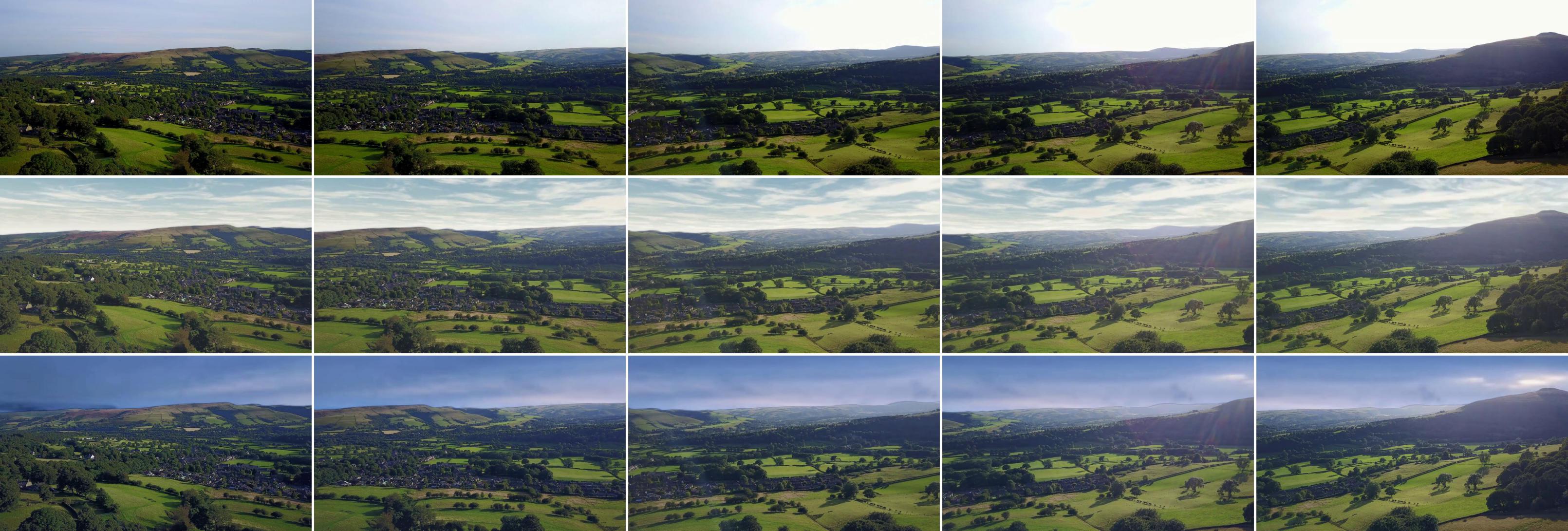}} \\  \\
(d) & {\includegraphics[width=0.9\linewidth]{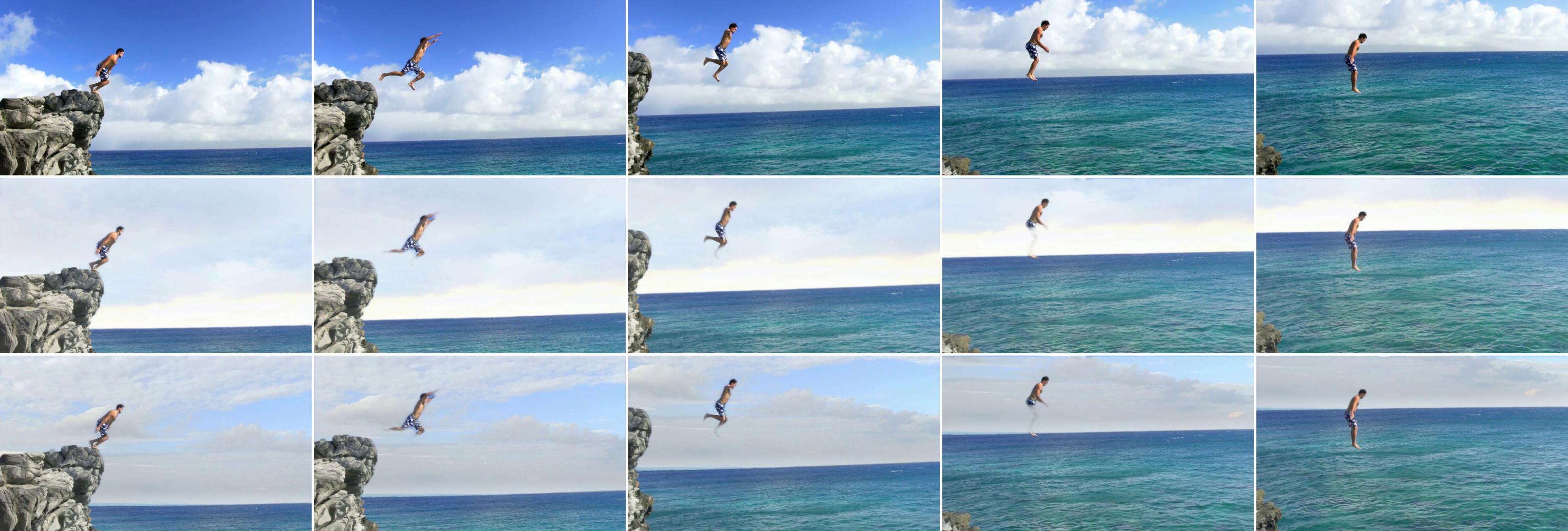}}
\end{tabular}
\caption{Additional results.}
\label{fig:results2}
\endgroup
\end{figure*}

\section{Conclusion}
We introduced an almost real time sky replacing framework for video,
adding a useful and powerful tool to the Augmented Reality toolbox.
Usually AR inserts objects close to the camera, where the geometry can be measured. We extended this  to insert content into areas which are essentially infinitely far away.

  

\bibliographystyle{eg-alpha-doi}

\bibliography{EGauthorGuidelines-conf-fin_with_teaser.bib}

\end{document}